\documentclass[10pt]{article} %
\usepackage[accepted]{tmlr}

\usepackage{amsmath,amsfonts,bm}

\def\eqref#1{equation~\ref{#1}}

\def\1{\bm{1}}

\DeclareMathAlphabet{\mathsfit}{\encodingdefault}{\sfdefault}{m}{sl}
\SetMathAlphabet{\mathsfit}{bold}{\encodingdefault}{\sfdefault}{bx}{n}

\newcommand{\R}{\mathbb{R}}

\usepackage{xspace}
\usepackage{url}            %
\usepackage{booktabs}       %
\usepackage{amsfonts}       %
\usepackage{nicefrac}       %
\usepackage{microtype}      %
\usepackage{xcolor}         %
\usepackage{epsfig}
\usepackage{graphicx}
\usepackage{amsmath}
\usepackage{amssymb}
\usepackage{wrapfig}
\usepackage{algorithm}
\usepackage{listings}
\usepackage{colortbl}
\usepackage{glossaries}
\usepackage{algorithm}
\usepackage{multirow}
\usepackage{listings}
\usepackage{fdsymbol}
\usepackage{enumitem}

\usepackage[normalem]{ulem}
\usepackage[font=small,skip=0pt]{caption}
\usepackage{pifont}%
\usepackage{framed}
\usepackage{color, colortbl}
\usepackage{subcaption,comment}
\usepackage{algorithm}
\usepackage{algpseudocode}
\definecolor{citecolor}{HTML}{0071bc}
\usepackage[pagebackref, breaklinks=true,bookmarks=false,colorlinks,bookmarks=false, citecolor=citecolor]{hyperref}

\definecolor{Gray}{gray}{0.9}

\newcommand{\myres}[1]{{$#1\!\times\!#1$}}   %

\def\fig#1{Figure~\ref{fig:#1}}

\def\sect#1{Section~\ref{sec:#1}}

\def\R{{\mathbb{R}}} %

\def\eg{\emph{e.g}\@\xspace}

\def\mypar#1{\vspace{0.01em}{\noindent\bf #1}\quad}

\newcommand{\ours}{\textsc{PQ-MIM}\xspace}

\newcommand{\tecnick}{\textsc{Tecnick}\xspace}

\newcommand{\clic}{\textsc{CLIC}\xspace}
\newcommand{\kodak}{\textsc{Kodak}\xspace}

\def\openreview{https://openreview.net/forum?id=Z2L5d9ay4B}
\def\month{03}
\def\year{23}

\definecolor{highlightRowColor}{rgb}{0.95, 0.95, 0.95}

\begin{document}

\title{Image Compression with \\ Product Quantized Masked Image Modeling}

\author{\scalebox{0.9}{\name Alaaeldin El-Nouby$^{\diamondsuit,\mathsection,\dagger}$} \email aelnouby@meta.com \\
        \scalebox{0.9}{\name Matthew Muckley$^{\diamondsuit}$} \email mmuckley@meta.com \\
        \scalebox{0.9}{\name Karen Ullrich$^{\diamondsuit}$} \email karenu@meta.com \\
        \scalebox{0.9}{\name Ivan Laptev$^{\mathsection,\dagger}$} \email ivan.laptev@inria.fr \\
        \scalebox{0.9}{\name Jakob Verbeek$^{\diamondsuit}$} \email jjverbeek@meta.com \\
        \scalebox{0.9}{\name Herv\'e J\'egou$^{\diamondsuit}$} \email rvj@meta.com \\ [0.1cm]
       \scalebox{0.9}{\addr $\diamondsuit$Meta AI, FAIR Team, Paris, $\mathsection$ENS, PSL University,  $\dagger$INRIA, Paris}
        }
\maketitle

\begin{abstract}

    Recent neural compression methods have been based on the popular hyperprior framework. It relies on Scalar Quantization and offers a very strong compression performance. This contrasts from recent advances in image generation and representation learning, where Vector Quantization is more commonly employed.  

In this work, we attempt to bring  these lines of research closer by revisiting vector quantization for image compression.
We build upon the VQ-VAE framework and introduce several modifications.
First, we replace the vanilla vector quantizer by a product quantizer. This intermediate solution between vector and scalar quantization  allows for a much wider set of rate-distortion points: It implicitly defines high-quality quantizers that would otherwise require intractably large codebooks. Second, inspired by the  success of Masked Image Modeling (MIM) in the context of self-supervised learning and generative image models, we propose a novel 
conditional entropy model which improves entropy coding by modelling the co-dependencies of the quantized latent codes.  The resulting \ours model is surprisingly effective: its compression performance on par with recent hyperprior methods. 
It also outperforms HiFiC in terms of FID and KID metrics when optimized with perceptual losses (e.g. adversarial). 
Finally, since \ours is compatible with image generation frameworks, we show qualitatively that it can operate under a hybrid mode between compression and generation, with no further training or finetuning. 
As a result, we explore the extreme compression regime where an image is compressed into 200 bytes, i.e., less than a tweet.

\end{abstract}

\section{Introduction}
\label{sec:intro}

Image compression played a crucial role in accelerating multiple events in history: it was a major component of the Voyager mission~\citep{ludwig2016voyager}. Efficient image codecs have accelerated the rapid growth of the internet by enabling the transmission of images in a few dozens of kilobytes, thanks to the emergence of effective lossy methods. This democratization was accompanied by standardization  efforts to facilitate the interoperability, which led to the emergence of standards such as the Joint Photographic Experts Groups (JPEG). Subsequent formats have leveraged scientific advances on all components of source coding, ranging from transforms~\citep{antonini1992image}, and quantization~\citep{gray98trinf}, to entropy coding~\citep{witten1987arithmetic,taubman2000high}, eventually leading to modern video compression codecs enabling streaming and video-conferencing applications.  

Neural methods have recently become increasingly popular for image compression as well as other image processing tasks, such as denoising~\citep{tian20nn}, super-resolution~\citep{bruna2015super,dong2015image,ledig2017photo,wang21pami} or image reconstruction~\citep{wang2020deep,knoll2020deep}. 
In typical scenarios, neural image compression is not necessarily mature enough to take over standard techniques like the BPG format inherited from the High-Efficiency Video Coding standard~\citep{sullivan2012overview}. 
This is because they do not offer a significant quantitative advantage over prior works that would justify the higher complexity, which depends on the context and operational constraints. 
A key advantage of neural compression methods is their enhanced qualitative reconstruction when 
incorporating an adversarial loss or likewise psycho-visual objectives favoring visually appealing reconstructions~\citep{agustsson19iccv, mentzer20nips}. 
From this perspective, neural compression is related to image generation. The two subfields, however, are currently dominated by different approaches, noticeably they employ different discretization procedures.  Indeed, while earlier neural compression methods utilized vector quantization \citep[VQ]{agustsson17nips}, recent methods mostly employ scalar quantization (SQ). 
In contrast, the recent literature on image generation \citep{chang22arxiv, yu21iclr, esser21cvpr, rombach2021highresolution} relies on Vector Quantization jointly with a distortion criterion akin to those used in compression.

In this work we aim to reduce the methodological gap and to make a step towards unification of neural image compression and image generation, and allowing image compression to more directly benefit from the rapid advances in image generation methods. Patch-based masking methods for self-supervised learning~\citep{bao21arxiv, he21mae, el2021large} have recently demonstrated their potential for image generation~\citep{chang22arxiv}. Inspired by this work we propose a compression approach built upon Vector Quantized Variational Auto-Encoders~\citep{oord17nips,razavi19nips}. In this context, we focus on two intertwined questions: (1)  How to define a vector quantizer offering a range of rate-distortion operating points? (2) How to define an entropy model minimizing the cost of storing the quantization indexes, while avoiding the prohibitive complexity of an auto-regressive model?

To address the challenges above, we revisit vector quantization in image compression, and investigate product quantization~\cite{jegou10pami} (PQ) in a compression system derived from VQ-VAE~\cite{oord16nips}. We show that PQ offers a strong and scalable rate-distortion trade-off. 
We then we focus on the spatial entropy modeling and coding of the quantization indexes in the VQ or PQ latent layer, hence we name our method as (Vector/Product)-Quantized Masked Image Modeling (VQ-MIM and PQ-MIM). To this end, we introduce a multi-stage vector-quantized image model: we gradually reduce the conditional entropy of the patch latent codes by increasing the number of observed patches we condition on for each stage. The conditional distribution over patches is estimated by a transformer model and provided to an entropy coder, symmetrically on the emitter and receiver sides. 

In summary, we make the following  contributions: 
\begin{itemize}
    \item We introduce a novel Masked Image Modeling conditional entropy model that significantly reduces the rates by leveraging the spatial inter-dependencies between latent codes.
    \item We introduce product quantization for VQ-VAE. This simple PQ-VAE variant offers a strong and scalable rate-distortion trade-off. 
    \item When trained with adversarial and perpetual losses, \ours exhibits a strong performance in terms of perceptual metrics like FID and KID, outperforming HiFiC \citep{mentzer20nips}.
    \item We qualitatively show that \ours is capable of operating in a hybrid mode, between generative and compression, without requiring further training and finetuning. This allows for higher resilience to corrupted or missing signal where our model can fill-in the missing information.
\end{itemize}

\begin{figure*}
    \centering
    \vspace{-9mm}
    \includegraphics[width=\linewidth]{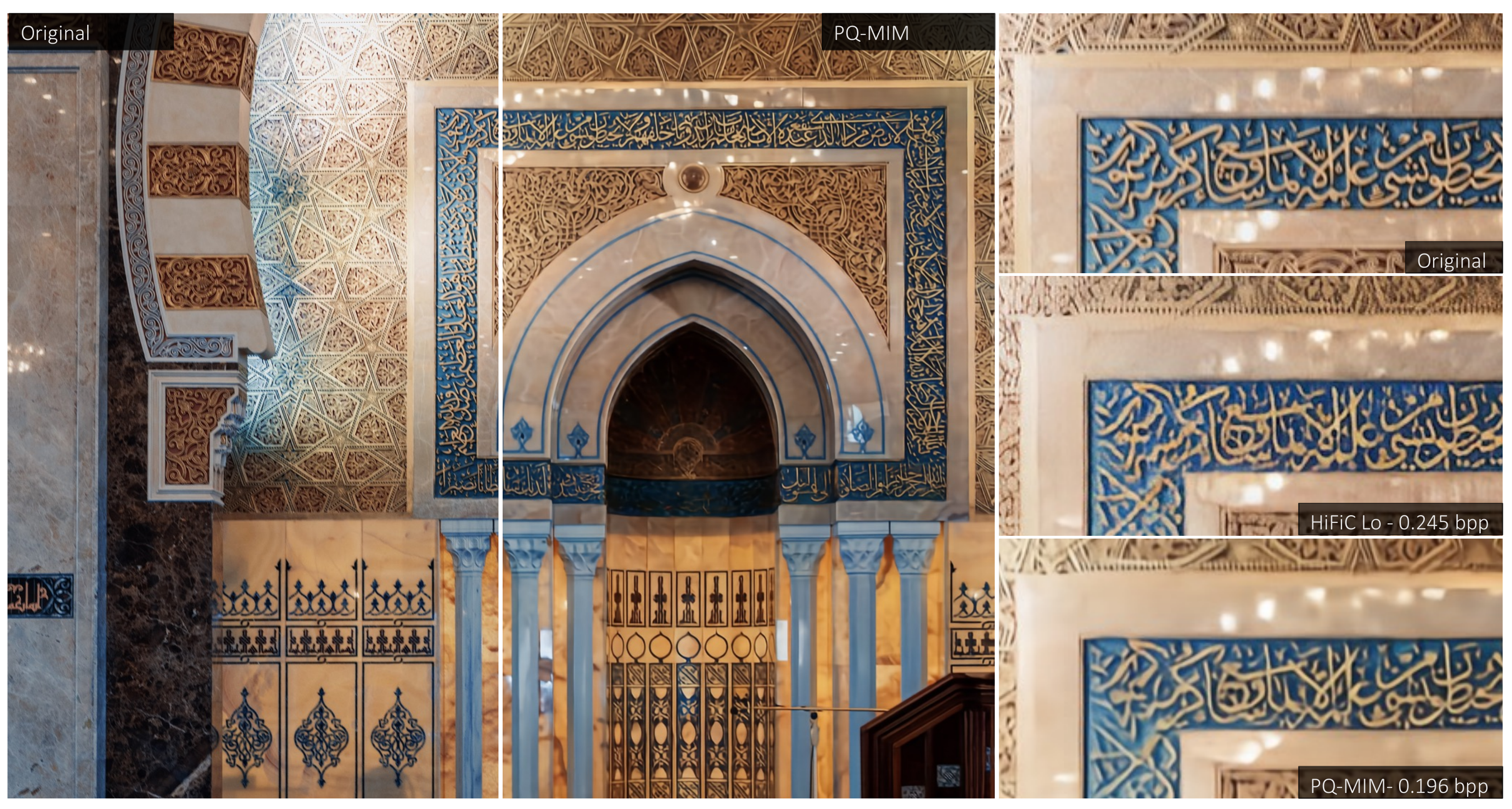}
    \caption{\textbf{Qualitative example of \ours compression.} \ours provides a strong compression performance. We retain many of the details present original image with minimal blurring effect even with comrpession rate as low as 0.196 bpp. Moreover, compared to HiFiC we achieve a lower rate for the same image. \ours provides colors that are more faithful to the original image, while HiFiC has a darkening effect and some high frequency artifacts. On the other hand, \ours can have some smoothing effect that can cause loss of detail for particular regions (e.g. some of the Arabic letters in the example above). More qualitative examples are provided in Appendix~\ref{app:qual_samples}.
    \label{fig:teaser}}
    \vspace{-4mm}
\end{figure*}
\section{Related work}
\label{sec:related}

\noindent \textbf{Neural image compression}
Early approaches to neural image compression reach back to the late 1980s \citep{sonehara1989image,sicuranza1990artificial,bottou1998high}. 
Recent  rapid advances in explicit and implicit density modelling~\citep{goodfellow14nips,kingma2013auto,larochelle2011neural,van2016wavenet,rezende2015variational,salimans17iclr} have renewed  interest in posing image compression as a learning problem. 
Due to the connection to variational learning~\citep{gregor2016towards,frey1998bayesian,alemi2018fixing}, variational auto-encoders have been the primary choice for lossy image compression. 
In contrast to standard variational models, the evaluation of neural compression models focuses on achievable bitrates, multi-scale applicability and computational complexity.
The initial works in the field used fully convolutional architectures for encoding/decoding~\citep{balle17iclr, theis2017lossy, mentzer18cvpr}.
The resulting encoded latent image representations are quantized and compressed via an entropy coder with learned explicit density model or ``entropy bottleneck''.
Initial variational approaches directly modeled a single level of code densities.
Ball\'e extended these models by introducing a second ``hyperprior''  that yielded improved performance~\citep{balle18iclr}.
Hyperprior models have been the basis for several subsequent advances with further improvements for density modeling, such as joint autoregressive models \citep{minnen18nips}, Gaussian mixture/attention \citep{cheng20cvpr}, and channel-wise auto-regressive models \citep{minnen2020channel}.
Another line of work has proposed to use vector quantization with histogram-based probabilities for image compression \citep{agustsson17nips,lu2019learning}.
Contrary to VQ-VAE models, these models typically optimize the rate (or a surrogate of the rate) directly and may include a spatial component for the quantized vectors.
Yang et al. \citep{yang2020improving} showed multiple ways to improve the encoding process, including the fine-tuning of the discretization process and employing bits-back coding in the entropy bottleneck.  
Finally, \citep{santurkar2018generative,mentzer20nips, rippel17icml, agustsson19iccv} showed that altering the distortion metric to include an additional adversarial loss can make a large difference for compression rate.
Another interesting line of work considers image compression by training image-specific networks, or network adapters, that map image coordinates to RGB values, and compressing  the image-specific parameters~\citep{dupont21arxiv,dupont22arxiv,strupler22eccv}.

\mypar{VQ-models for image generation.}
There has been significant interest in generative models based in discrete image representations, as introduced by VQ-VAE~\citep{oord17nips,razavi19nips}.
A discrete representation of reduced spatial resolution is learned by means of an autoencoder which quantizes the latent representations.
This discrete representation is coupled with a strong prior, for example implemented as an autoregressive pixel-CNN model. 
VQ-GAN~\citep{esser21cvpr} replaces the prior architecture with a transformer model~\citep{vaswani17nips}, and introduces  an adversarial loss term to learn an autoencoder with more visually pleasing reconstructions and improved sample quality.
This approach has  been extended to text-based generative image models by extending the prior to model a longer sequence that combines the discrete image representation with a prefix that encoding the conditioning text.
This has yielded impressive results by scaling the model capacity and training data to tens or hundreds of million text-image pairs~\citep{ding21arxiv,gafni22arxiv,ramesh21arxiv}. 
A  fundamental limitation of   autoregressive  generative models is that they sample data sequentially,  requiring separate non-parallel evaluation of the predictive (transformer) model to sample each token.
To alleviate this, several data items can be sampled independently in parallel, conditioning on all previously sampled tokens. 
This has been leveraged to speed-up pixel-CNNs for small images and video by two to three orders of magnitude~\citep{reed17icml}.
More recently, MaskGIT~\citep{chang22arxiv} and follow-up work~\citep{lezama22eccv} explored this for generative models of VQ-VAE representations, and find that similar or better sample quality is obtained by parallel sampling of image patch subsets in few steps, reducing the generation time significantly.
Despite their success for image synthesis we are not aware of the use of such models for image compression in earlier work.

\section{Product Quantized Masked Image Modeling}
\label{sec:method}

\begin{figure*}[t!]
    \centering
    \vspace{-9mm}
    \includegraphics[width=0.95\linewidth]{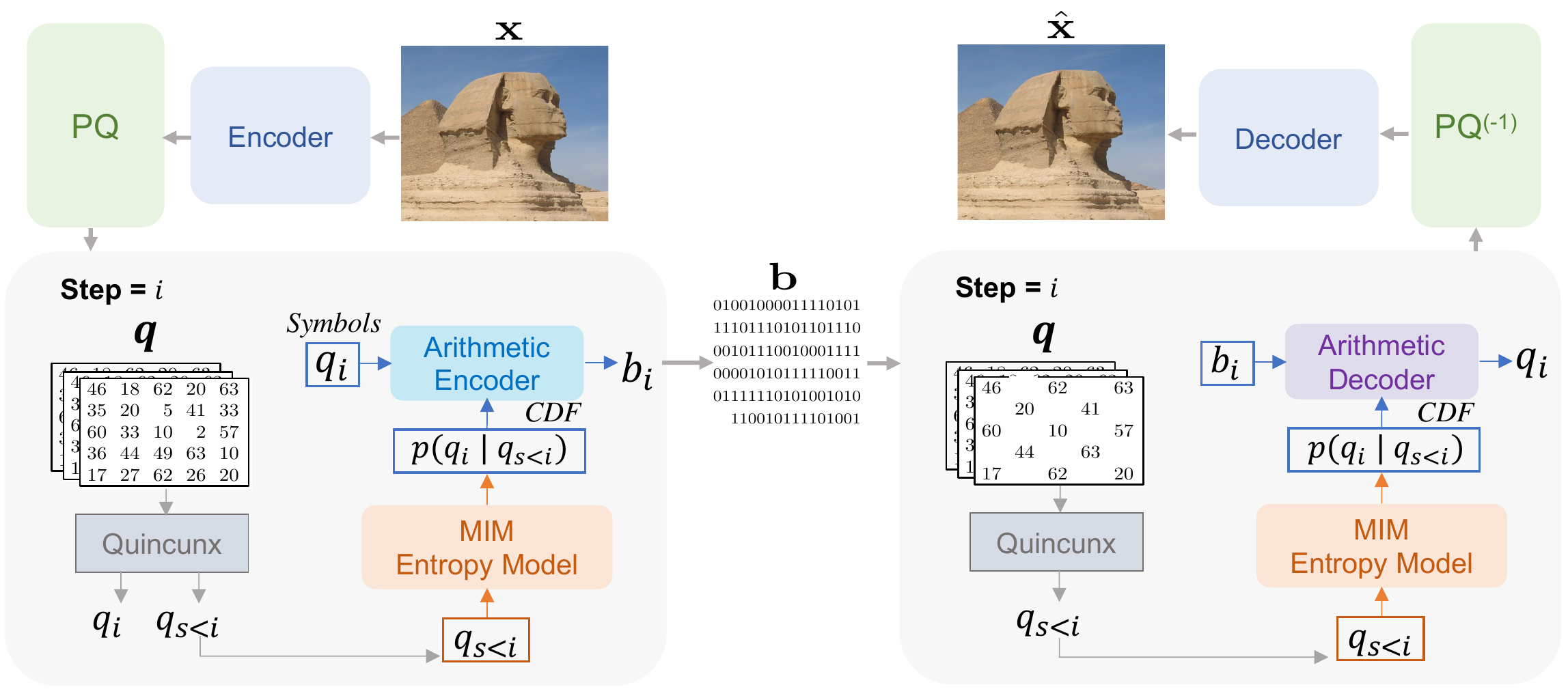}
    \caption{\textbf{\ours overview.} Our model consists of (i) a  a transformer based encoder and decoder, (ii) a masked image model (MIM) for conditional entropy modeling, and (iii) an entropy coder, \eg an arithmetic coder (AE/AD).
    The input image $\mathbf{x}$ is projected to a set of latent features, followed by product
quantization to yield quantization indices $\mathbf{q}$. The arithmetic coder encodes (and decodes) $\mathbf{q}$ into a bitstream $\mathbf{b}$ in a lossless manner.
    The elements in $\mathbf{q}$ are spatially split into groups, as detailed in Figure~\ref{fig:spatial_model}. 
    \textbf{Conditional Entropy Modeling.} Our model estimates the conditional probabilities of the discrete indices in $\mathcal{S}$ steps. Every step, a subset of the tokens $q_i$ is selected using the quincunx pattern. Our MIM transformer estimates $p(q_i | q_{s < i})$ and passes it to the Arithmetic encoder as a CDF, effectively reducing the lossless compression cost. 
    }
    \label{fig:overview}
    \vspace{-3mm}
\end{figure*}

This work takes a step towards to closing the gap between neural image compression and image generation methodologies. We revisit vector quantization for image compression and propose an entropy model inspired by masked image modelling. Our compression pipeline, depicted in \fig{overview}, relies on three neural networks:
\begin{enumerate}
\item \textbf{The Encoder network} $\hat{E}: \mathcal{X} \rightarrow \mathcal{Z} $ maps input images $\mathbf{x} \in \mathcal{X}$ to a quantized representation $\mathbf{z} \in \mathcal{Z}$. 
\item \textbf{The Mask Image Model} (MIM) compresses the quantized representations without loss of information. This network is involved both on the compression and decompression side. 
\item \textbf{The Decoder network} $G: \mathcal{Z} \rightarrow \mathcal{X}$ produces an estimate $\hat{\mathbf{x}}=G(\hat{E}(\mathbf{x}))$ of the original image $\mathbf{x}$. 
\end{enumerate}

We now detail the architecture, in particular our PQ proposal, of the image model that we employ in the statistical lossless coding, as well as the training scheme.  

\subsection{High-level architecture: PQ-VAE}

\paragraph{High-level architecture.} We follow recent work on discrete generative image models for the design of image encoder and decoder~\citep{chang22arxiv,esser21cvpr,oord17nips,razavi19nips,yu21iclr}.
The encoder $E: \mathcal{X}\rightarrow \R^{w\times h\times d}$ takes an RGB image $\mathbf{x}$ of resolution $W\!\times\!H$ as input and  maps it to a latent representation $E(\mathbf{x})$ with $d$ feature channels and a reduced spatial resolution $w\!\times\!h$,  downsampling the input resolution by a factor $f={W}/{w}={H}/{h}$. 
The $T=w\!\times\!h$ elements of the latent representation $E(\mathbf{x})$ are quantized with a vector quantizer $Q(\cdot)$ to  produce the quantized latent representation  $\mathbf{z}=Q(E(\mathbf{x}))=\hat{E}(\mathbf{x})$, where each element in  $E(\mathbf{x})$ is replaced with its nearest cluster center.
The decoder $G$ uses the quantized latents $\mathbf{z}$  to reconstruct the image.

\mypar{Product Quantization.}
In VQ-VAE, the quantizer $Q$ is simply an online k-means quantizer that produces quantization indices from real-valued vectors. 
We denote by $\mathbf{q}\in\{1,\dots,V\}^T$ the map of quantization indices indicating for each element of $\mathbf{z}$ which of the $V$ centroids is selected. 
The higher $V$ the more precise is the approximation $\mathbf{z}$, leading to higher bit rates. For instance, assuming that indices are coded with a naive coding scheme (see next section), the bit rate is doubled when moving to $V=256$ to $V=65536$ centroids. 

\begin{wrapfigure}{r}{0.45\textwidth}
  \begin{center}
  \vspace{-1mm}
    \includegraphics[trim={0.0cm, 1.0cm, 1.0cm, 0.0cm}, clip, width=1\linewidth]{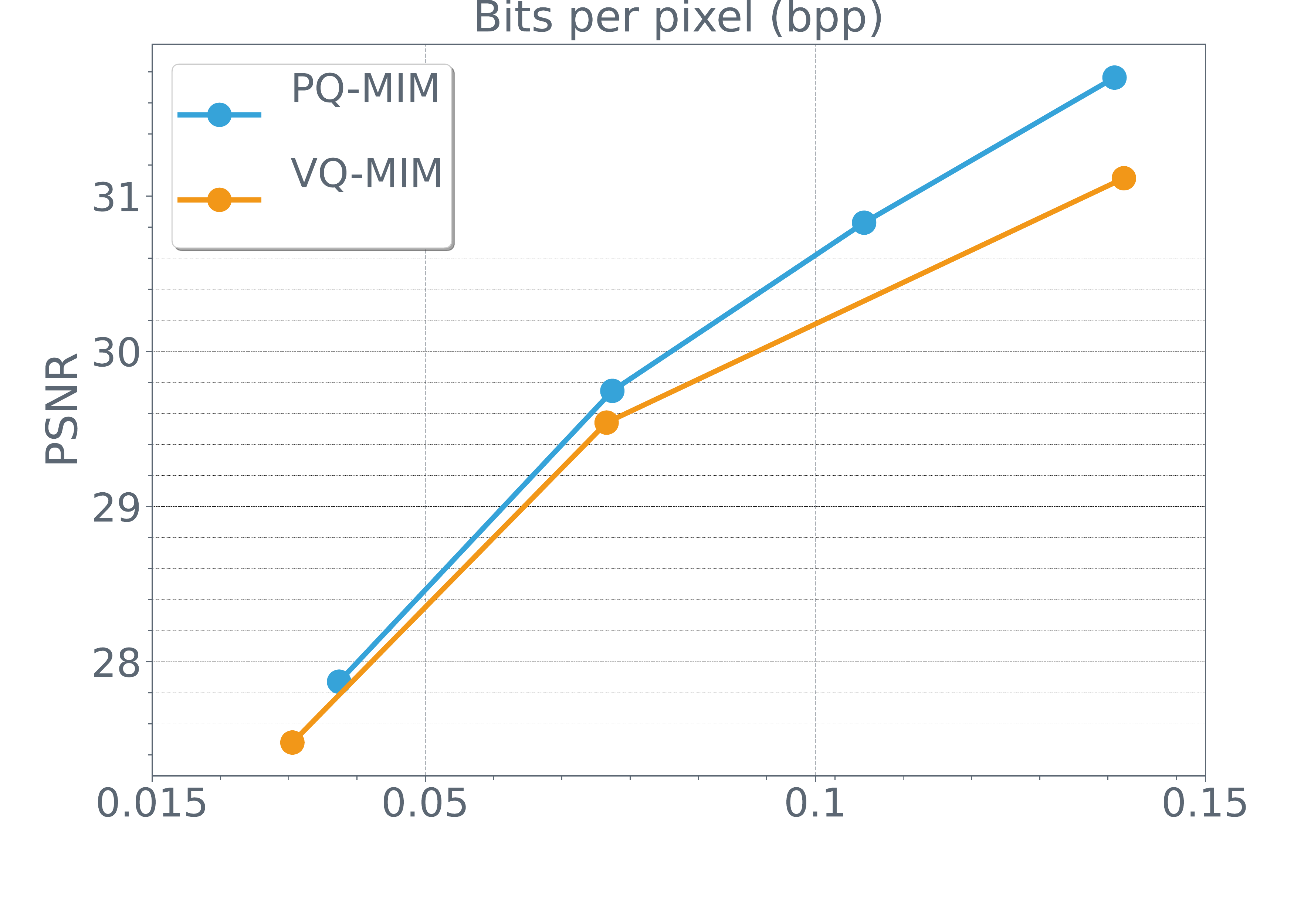}
    \caption{\textbf{PQ and VQ comparison}. While VQ provides a comparable performance to PQ for extremely low rates where the codebook size is small, PQ exhibits better scaling behaviour for higher rates.  }
    \label{fig:pq_vs_vq}
    \end{center}
    \vspace{-5mm}
\end{wrapfigure}

However, scaling the number of centroids $K$ is possible up to thousands of centroids, but beyond that it is computationally prohibitive. Additionally, it is challenging to train large codebooks where each centroid has a very low probability of being updated. %
To address this problem, we replace the online k-means quantizer by a product quantizer~\citep{jegou10pami} (PQ): the latent vector $\mathbf{z}$ is split into $M$ subvectors as $\mathbf{z}=[\mathbf{z}^1,\dots,\mathbf{z}^j,\dots,\mathbf{z}^M]$ of dimension $M/d$. Each subvector is quantized by a distinct quantizer having $V_\mathrm{s}$ quantization value. The set of quantizers implicitly defines a vector quantizer in the latent space with $V=V_\mathrm{s}^M$ distinct centroids. Hence, we can easily define very large codebooks without the computational and optimization problems mentioned above, because both the assignment and learning are marginalized over the different subspaces. Empirically, we observe in Figure~\ref{fig:pq_vs_vq} that PQ provides a better scaling behaviour for higher rates compared to VQ whose codebook size needs to grow exponentially to achieve the same rates.

\mypar{Neural network.}
Without loss of generality, we choose all neural network models to be identical. 
This is not a requirement but this offers the property that the encoder and decoder have identical complexities, and that the memory and compute peaks are identical.  
More specifically we choose a cross-variance transformer~\citep{elnouby21nips} (XCiT), whose complexity is linear with respect to image resolution. In contrast, standard vision transformers~\citep{dosovitskiy21iclr} (ViT) are quadratic in the image surface, which is prohibitive for high resolution images that can typically require strong compression. We point out that recent work has shown that Swin-Transformers~\citep{zhu2021transformer} could be a compelling choice as well in the context of image compression. Formally they also have a quadratic complexity, but this is amortized by the hierarchical structure of this architecture.

\subsection{Image entropy model}

In this section we present PQ-MIM. 
The objective is to compress the discrete representations $\bm q$ without loss of information, producing a bitstream of the compressed representation that can be transmitted or stored. 
During the decoding stage, we invert the aforementioned lossless compression. %
This model can be regarded as the VQ/PQ-VAE counterpart of adaptive contextual arithmetic coders, like EBCOT~\citep{taubman2000high} or CABAC~\citep{richardson2004h}, proposed in early compression standards, in that it couples a conditional probabilistic model with an arithmetic coder.

\begin{figure*}
    \centering
    \vspace{-2mm}
    \includegraphics[width=\linewidth]{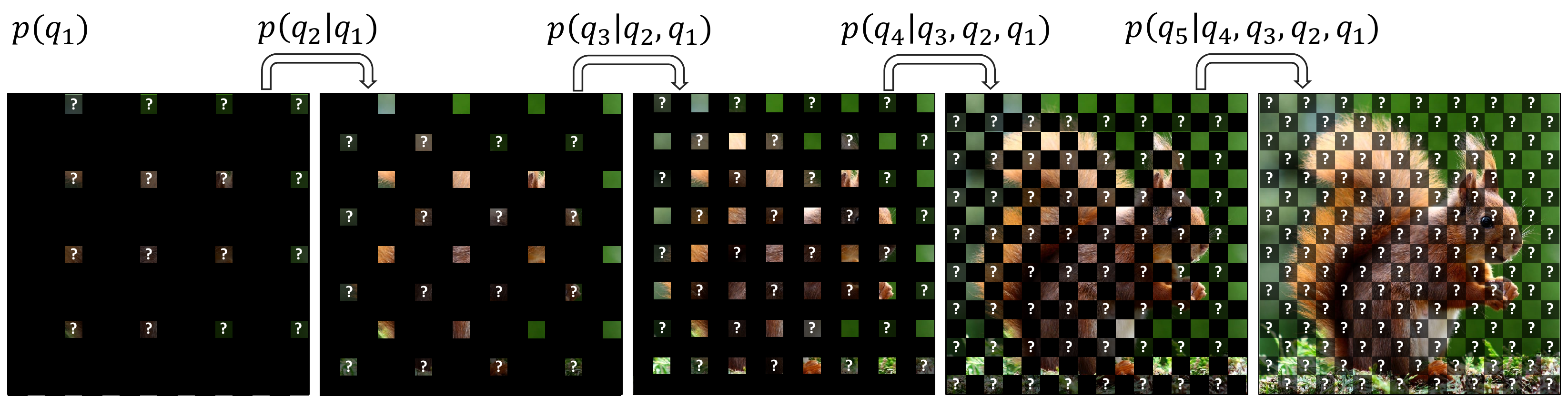}
    \caption{\textbf{Illustration of PQ-MIM with a quincunx pattern.} We employe the quincunx pattern both on the encoder and decoder side. 
    Each panel represents one of five stages in which we en/de-code a set of tokens in parallel using a probability model $p_s$  implemented by a transformer with parameters $\theta_s$.
    The transformer predicts the tokens in $\mathbf{q}_s$, displayed in grayscale and marked by ``\textbf{?}'', and takes as input the preceding groups of tokens $\mathbf{q}_1,\dots,\mathbf{q}_{s-1}$ that are displayed in color. %
    The distribution provided by this neural network is fed to an arithmetic en/de-coder. %
    \label{fig:spatial_model}
    }
    \vspace{-3mm}
\end{figure*}

\mypar{Lossless compression.} 
A naive manner for lossless compression of the  discrete image codes $\mathbf{q}=\{q_t\}_{t=1}^T$ is to use fixed-length codes.  
In that case, each code word is assigned to a unique binary representation of equal length, resulting in $\lceil{\log_2 V}\rceil$ bits per element $q_t$. 
This approach is computationally very efficient as  fixed-length codes are not model-based, and as such  do not require likelihood estimation, and  because all codes are of equal length by construction, the computation is perfectly parallelizable. 
However, theoretically this coding scheme could be Shannon optimal only if codewords are uniformly and independently distributed. 
Those assumptions are not met in practice due to the architecture choices we have made previously: k-means does not produce uniformly distributed indices except in singular cases~\citep{gray98trinf}. More details about losseless compression are covered in Appendix~\ref{app:lossless}.

\paragraph{Entropy model.}
To improve the bitrate, we hence rely on an entropy coder, which provides an inverse pair of functions, $\texttt{enc}_{p}$ and $\texttt{dec}_{p}$, achieving near optimal compression rates on sequences of %
symbols for any distribution $p$. 
The better $p$ matches the (unknown) underlying data distribution, the better the compression rate~\citep{cover2012elements}. Generally, more powerful generative models will ensure better compression performance. %

Fully autoregressive generative models $p(\mathbf{q})=\prod_{t=1}^T p(q_t|\mathbf{q}_{<t})$ are  powerful ~\citep{ding21arxiv,esser21cvpr,gafni22arxiv,ramesh21arxiv,yu21iclr}, 
however, they are inconvenient in that the likelihood estimation for this type of model is not trivially parallelizable: each patch index must be processed sequentially as it is used to condition subsequent patch indices. 
Thus, similar to prior works~\citep{chang22arxiv,reed17icml} we propose a masked image model, 
  which we use to predict the image patch indices in several stages. %
Specifically, we partition $\mathbf{q}$ into $S$ subsets $\mathbf{q}_1, \mathbf{q}_2, \dots, \mathbf{q}_S$ of patch indices, that we refer to as \emph{tokens} by analogy to language modelling:
\begin{equation}
\mathbf{q} = \bigcup_{s=1}^S \mathbf{q}_s.  
\end{equation}
We model the elements in each subset conditionally independent given all preceding groups:
\begin{align}
    p(\mathbf{q}) & = \ \ \prod_{s=1}^{S} p\left(\mathbf{q}_s|\mathbf{q}_{<s};\theta_s\right),\\
    p\left(\mathbf{q}_s|\mathbf{q}_{<s};\theta_s\right) & = \prod_{q_t\in \mathbf{q}_s} p\left(q_t|\mathbf{q}_1, \mathbf{q}_2, \dots, \mathbf{q}_{s-1}; \theta_s\right).
\end{align}
Since $p(\mathbf{q}_1)$ is not conditioned on any previous elements, it fully factorizes over $q_t\in\mathbf{q}_1$, and  we model it as the marginal distribution over the  vocabulary observed on the training data.
The non-trivial conditional distributions $p\left(\mathbf{q}_s|\mathbf{q}_{<s};\theta_s\right)$ for $s\geq 2$ are modeled using  transformer  networks, which have few inductive biases and have been successful across many tasks, including image generation. An overview of the MIM entropy model is  illustrated in Figure~\ref{fig:overview}.

\mypar{Amortized encoding and decoding.} 
From a computational point of view, our proposal allows for the compression (resp. decompression) to proceed in $S$ stages. In each stage  $s$ we encode/decode the set $\mathbf{q}_s$ of tokens conditioned on the groups $\mathbf{q}_1, \dots, \mathbf{q}_{s-1}$ encoded/decoded in preceding stages, but independently among the tokens in the set $\mathbf{q}_s$. 
This allows for paralellization among the elements in each subset $\mathrm{q}_s$, and requires strictly $S$  forward passes through the model independent of the image size, rather than $T$ sequential forwards passes for fully autoregressive models.

\mypar{Quincunx partitioning.} 
In practice, we typically use $S=5$ stages. 
We have explored different patterns to partition the $T$ tokens over the $S$ stages. 
In particular we consider the ``quincunx'' regular grid pattern, where in each stage we double the number of tokens to predict, see \fig{spatial_model} for an illustration. This multi-level refinement was previously explored for image compression in the context of lifting schemes designed with oriented wavelets~\citep{chappelier2006oriented}. 
In our experiments we contrast this partitioning with alternative ones with other patterns and subset cardinalities (Figure~\ref{fig:mim_steps}).

\begin{figure}[t!]
\vspace{-5mm}
\begin{minipage}{\linewidth}
    \begin{subfigure}{0.48\linewidth}
        \centering
        \caption{\tecnick PSNR}
        \includegraphics[trim={0.0cm, 1.0cm, 1.0cm, 0.0cm}, clip, width=1.0\linewidth]{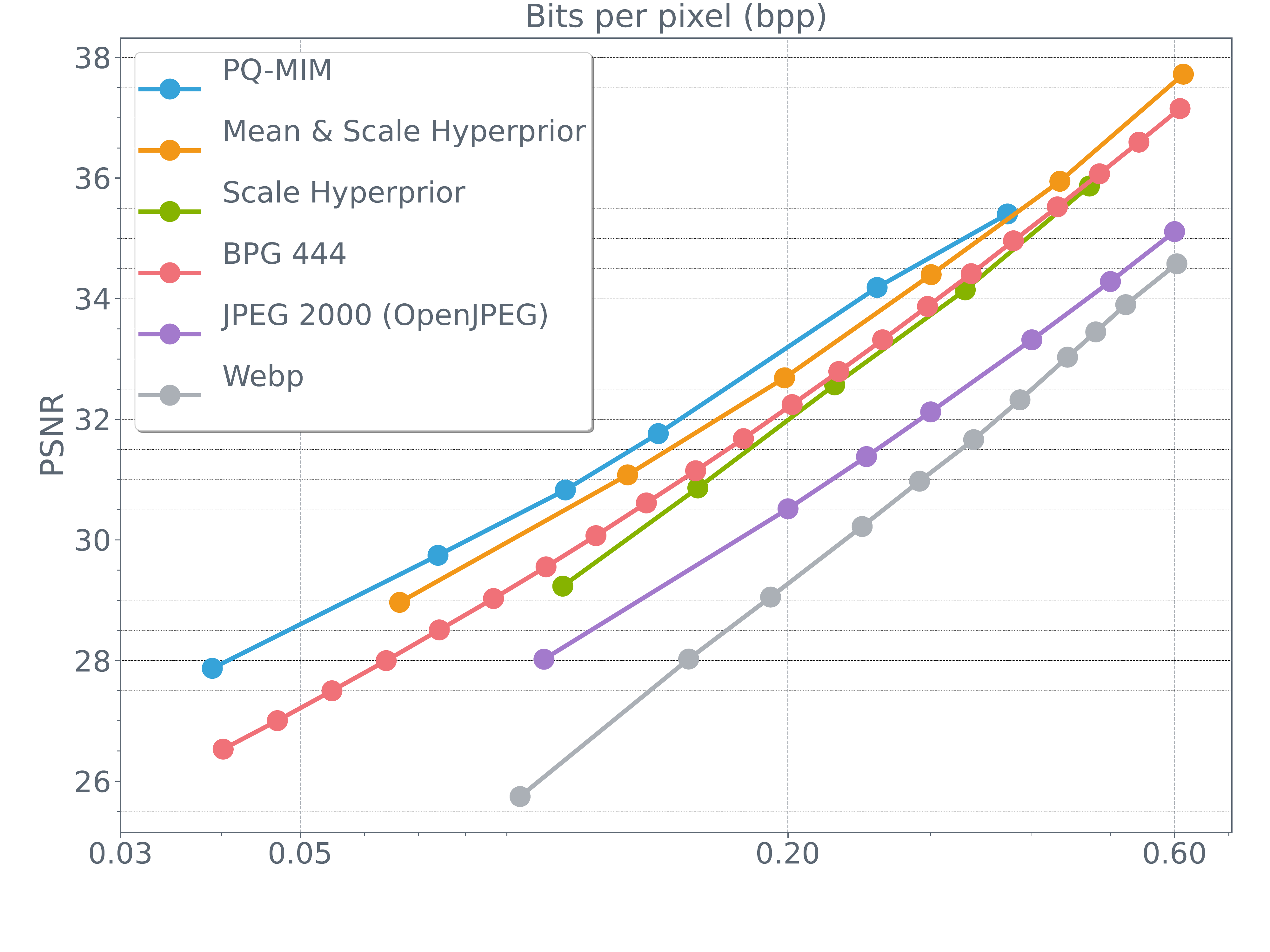}
    \end{subfigure}%
    \hfill
    \begin{subfigure}{0.48\linewidth}
        \centering
        \caption{\tecnick MS-SSIM}
        \includegraphics[trim={0.0cm, 1.0cm, 1.0cm, 0.0cm}, clip, width=1.0\linewidth]{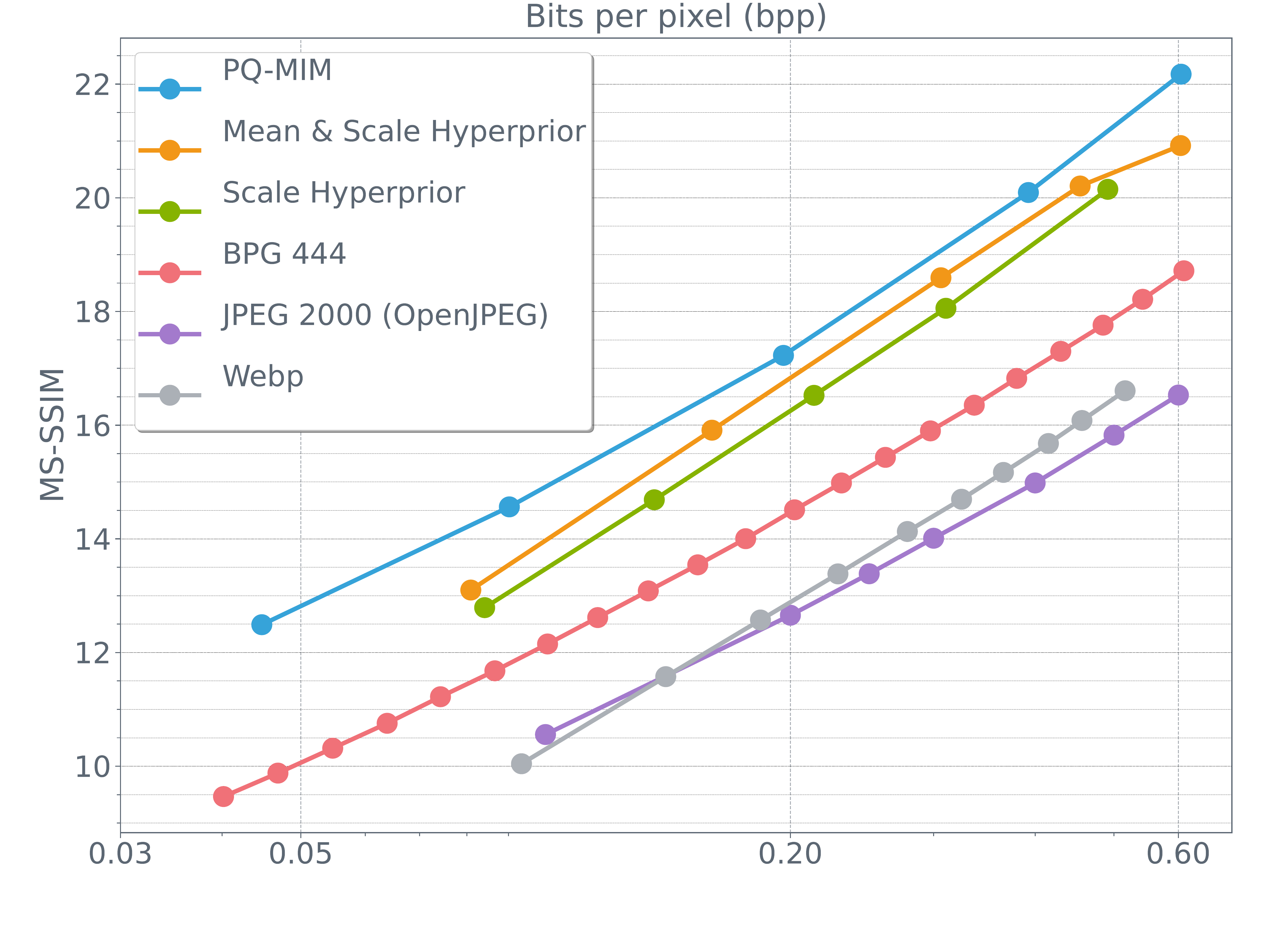}
    \end{subfigure}
\end{minipage}%
\hfill
\begin{minipage}{\linewidth}
    
\begin{subfigure}{0.48\linewidth}
        \centering
        \caption{\kodak PSNR}
        \includegraphics[trim={0.0cm, 1.0cm, 1.0cm, 0.0cm}, clip, width=1.0\linewidth]{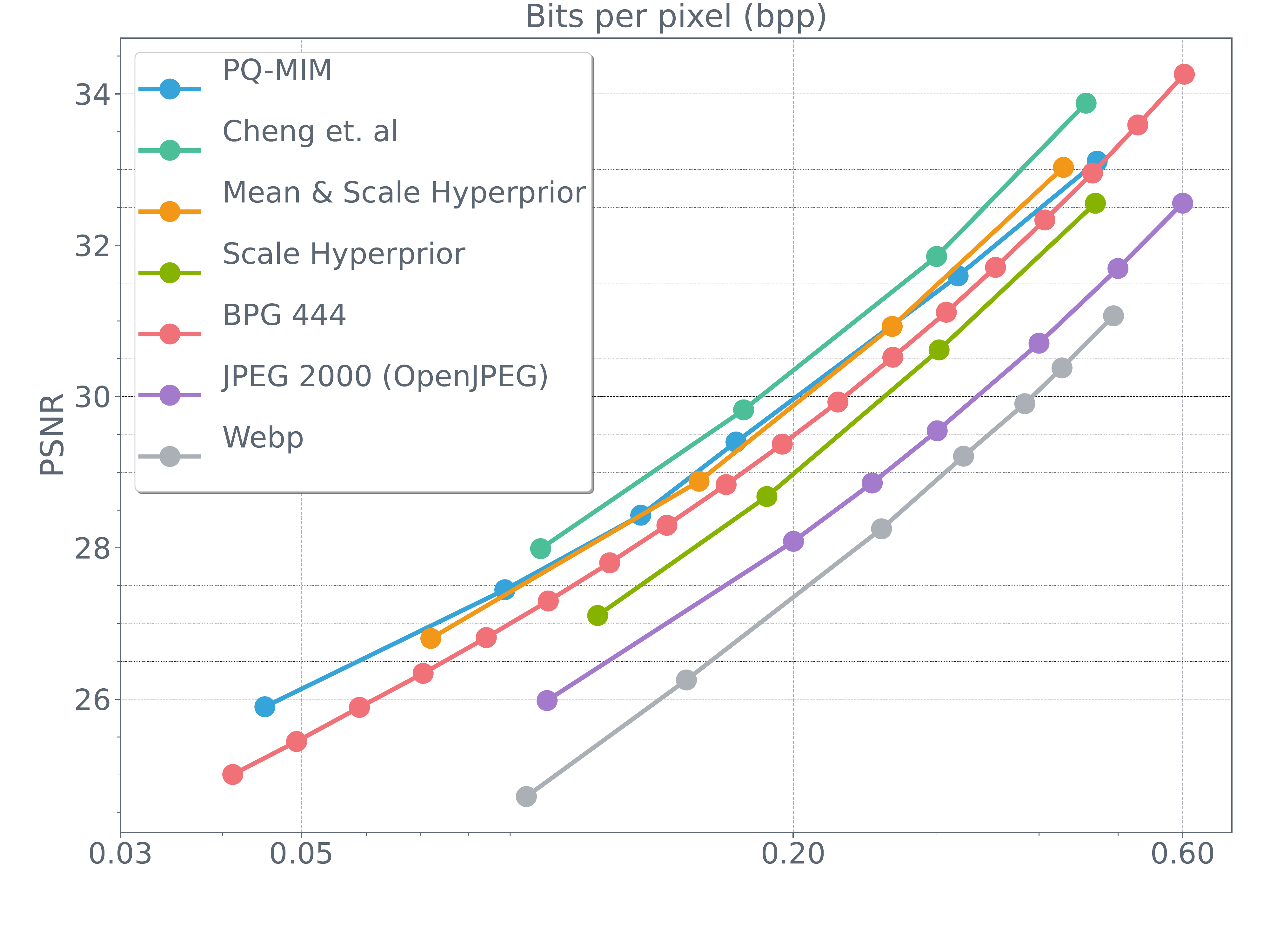}
    \end{subfigure}%
    \hfill
    \begin{subfigure}{0.48\linewidth}
        \centering
        \caption{\kodak MS-SSIM}
        \includegraphics[trim={0.0cm, 1.0cm, 1.0cm, 0.0cm}, clip, width=1.0\linewidth]{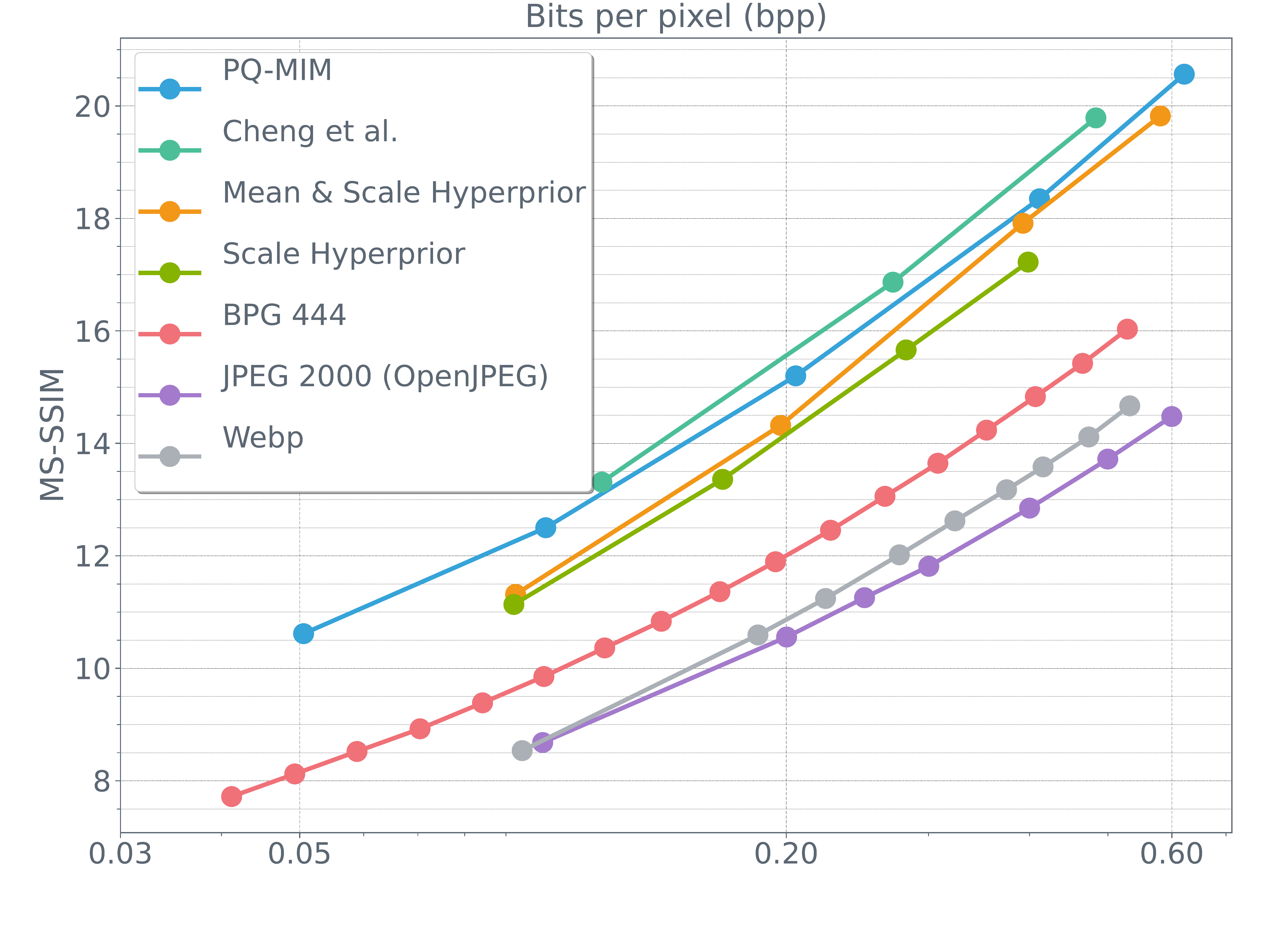}
    \end{subfigure}
    \caption{\textbf{Rate-Distortion performance for \tecnick and \kodak datasets.} We report \ours PSNR and MS-SSIM performance for various operating points. \ours provides a competitive performance, particularly for MS-SSIM, compared to standard codecs such as JPEG 2000 \citep{Taubman2012} and BPG \citep{ballard_bpg} as well as recent neural methods \citep{cheng20cvpr, minnen18nips, balle18iclr}.
}
\label{fig:results_psnr_msssim}
\end{minipage}
\end{figure}

\subsection{Training the \ours}

\mypar{Reconstruction objective and training.}
The goal of lossy image compression is to match the image and its reconstruction as closely as possible according to some distortion metric. In this paper, we train our model to reduce the image distortion and the quantization loss using the following objective:
\begin{equation}
    L = L_\textrm{Rec} + \eta \cdot L_\textrm{PQ}
\end{equation}

Following VQ-VAE, our quantization objective $L_\textrm{PQ}$ consists of an embedding and commitment losses, averaged over the $M$ different PQ sub-vectors. For the distortion loss $L_\textrm{Rec}$, we present two setups where we use different types of distortion measures:
 \begin{itemize}[leftmargin=*]
    \item \textbf{MSE \& MS-SSIM.} Typical distortion measures used in the majority of the neural compression literature such as mean squared error (MSE) or multi-scale structural similarity~\citep{wang2003multiscale} (MS-SSIM). For this setup, the model is  trained solely using one distortion measure at a time.
    \begin{equation}
    L_\textrm{Rec}(\mathbf{x},\hat{E},G) = L_\textrm{MSE/MS-SSIM}(\mathbf{x},\hat{\mathbf{x}})
\end{equation}
    \item \textbf{Perceptual measures.} Alternatively, we report a setup where we utilize perceptual objectives such as LPIPS~\citep{zhang18cvpr} and adversarial training~\citep{goodfellow14nips} to enhance psycho-visual image quality. The distortion loss is defined as:
    \begin{equation}
    L_\textrm{Rec}(\mathbf{x},\hat{E},G) = L_\textrm{MSE}(\mathbf{x},\hat{\mathbf{x}}) + \alpha \cdot L_\textrm{Perc}(\mathbf{x},\hat{\mathbf{x}}) + \gamma \cdot L_\textrm{Adv}(\hat{E},G,D), 
\end{equation}
where  $\alpha$ and $\gamma$ are weighing coefficients and the adversarial loss $L_\textrm{Adv}$ is defined as:
\begin{equation}
L_\textrm{Adv}(\hat{E},G,D) = \mathbb{E}_{\mathbf{x}}[ \ln D(\mathbf{x}) ] + \mathbb{E}_{\hat{\mathbf{x}}}[ \ln(1- D(\hat{\mathbf{x}})) ],
\label{eq:Ladv}
\end{equation}
where $D(\cdot)$ is the discriminator and $\mathbb{E}_\mathbf{x}$ denotes the expectation over $\mathbf{x}$ sampled uniformly from the training set. Similarly $\mathbb{E}_{\hat{\mathbf{x}}}$ denotes the expectation over reconstructed training images. 
Note that, unlike the MSE and perceptual losses, the adversarial loss does not compare individual images and their reconstructions, but aims to match the \emph{distributions} of original images and their reconstructions.

\end{itemize}

\mypar{Training the entropy model.} 
Our MIM module is an XCiT transformer that accepts $T$ tokens as input representing the image patches. During training, we randomly mask a set of tokens by sampling from a uniform distribution $U(0, 1)$. The masked tokens are replaced with a mask embedding vector, while the observed token indices are mapped to their corresponding continuous representation using an embedding look-up table. The MIM module outputs a context vector for every masked token which in turn is passed to $M$ linear heads, representing the different PQ sub-vector indices, followed by a softmax to yield a distribution $p({\bm q}_s| {\bm q}_{<s})$. 
The module is trained using a standard cross-entropy objective. While we train the autoencoder and the MIM modules simultaneously, we do not backpropagate gradients from the MIM module to the encoder $E$ or the quantization parameters, so the two components can be trained separately in sequence.

\section{Experiments}
\label{sec:exp}

\begin{figure*}
    \vspace{-5mm}
    \begin{subfigure}[t]{0.32\linewidth}
        \caption{FID}
        \includegraphics[trim={0.0cm, 2.0cm, 1.2cm, 0.0cm}, clip, width=1.0\linewidth]{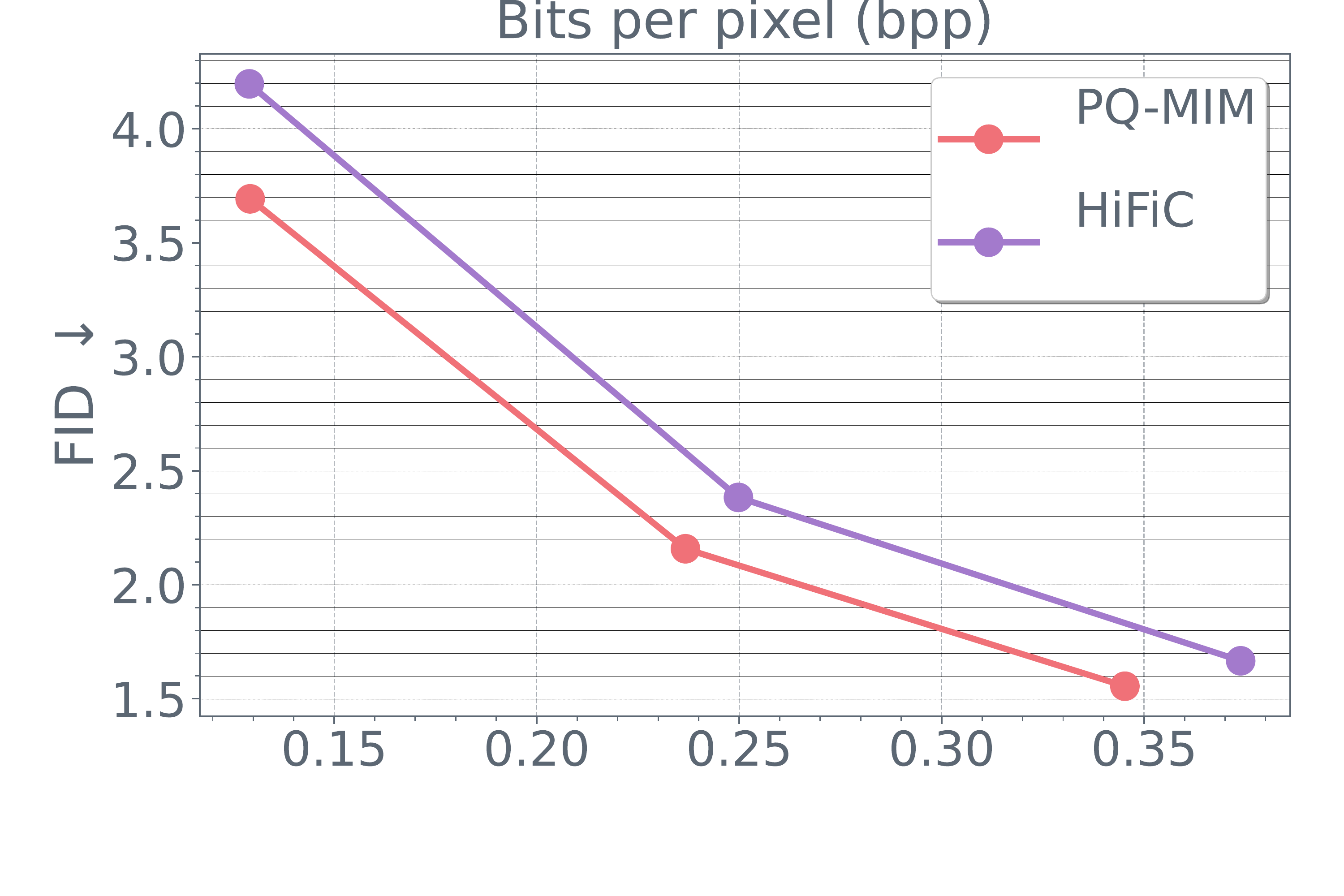}

    \end{subfigure}
    \hfill
    \begin{subfigure}[t]{0.32\linewidth}
        \centering
        \caption{KID}
        \includegraphics[trim={0.0cm, 2.0cm, 1.2cm, 0.0cm}, clip, width=1.04\linewidth]{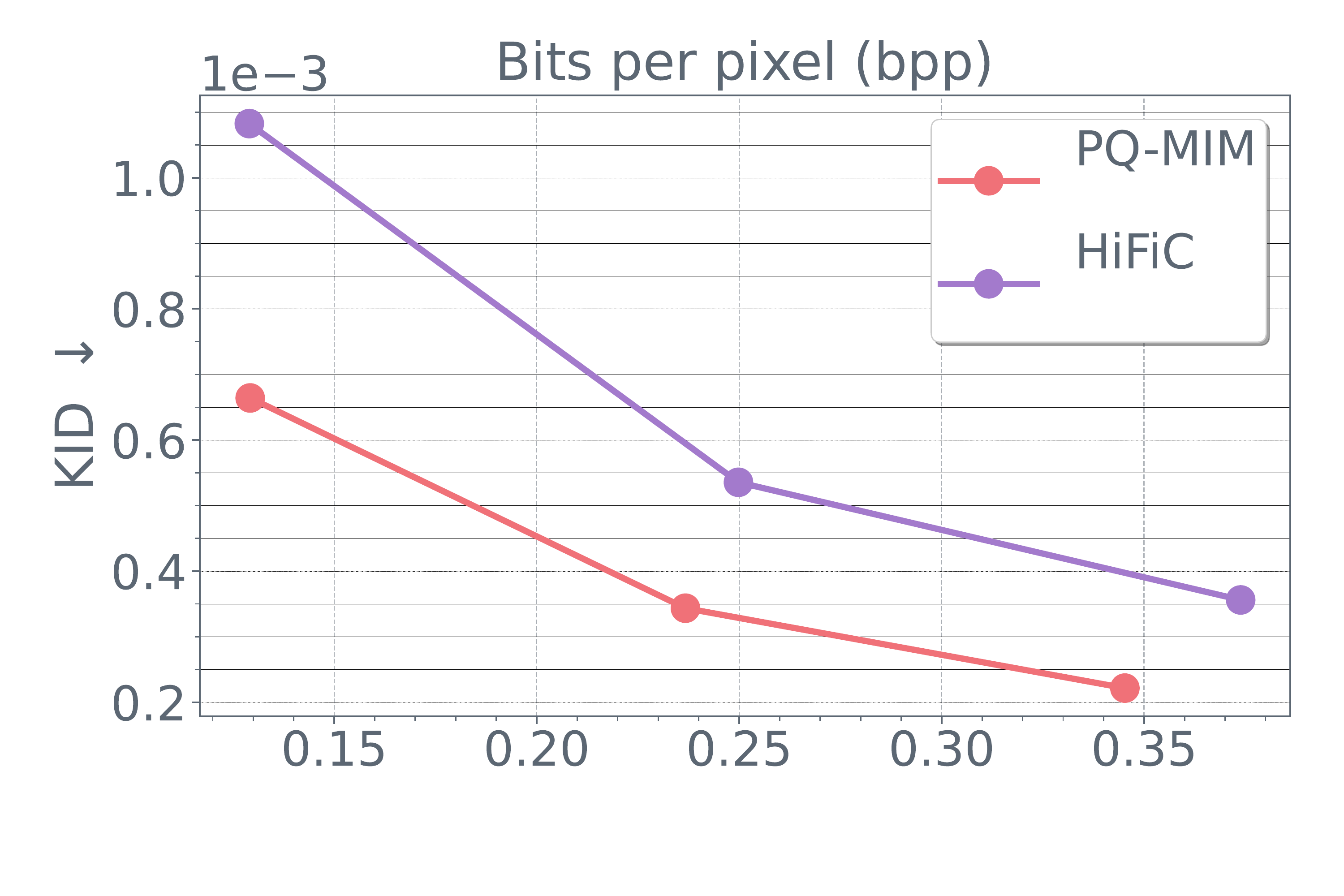}
    \end{subfigure}
    \hfill
    \begin{subfigure}[t]{0.32\linewidth}
        \centering
        \caption{MS-SSIM}
        \includegraphics[trim={0.0cm, 2.0cm, 1.2cm, 0.0cm}, clip, width=1.0\linewidth]{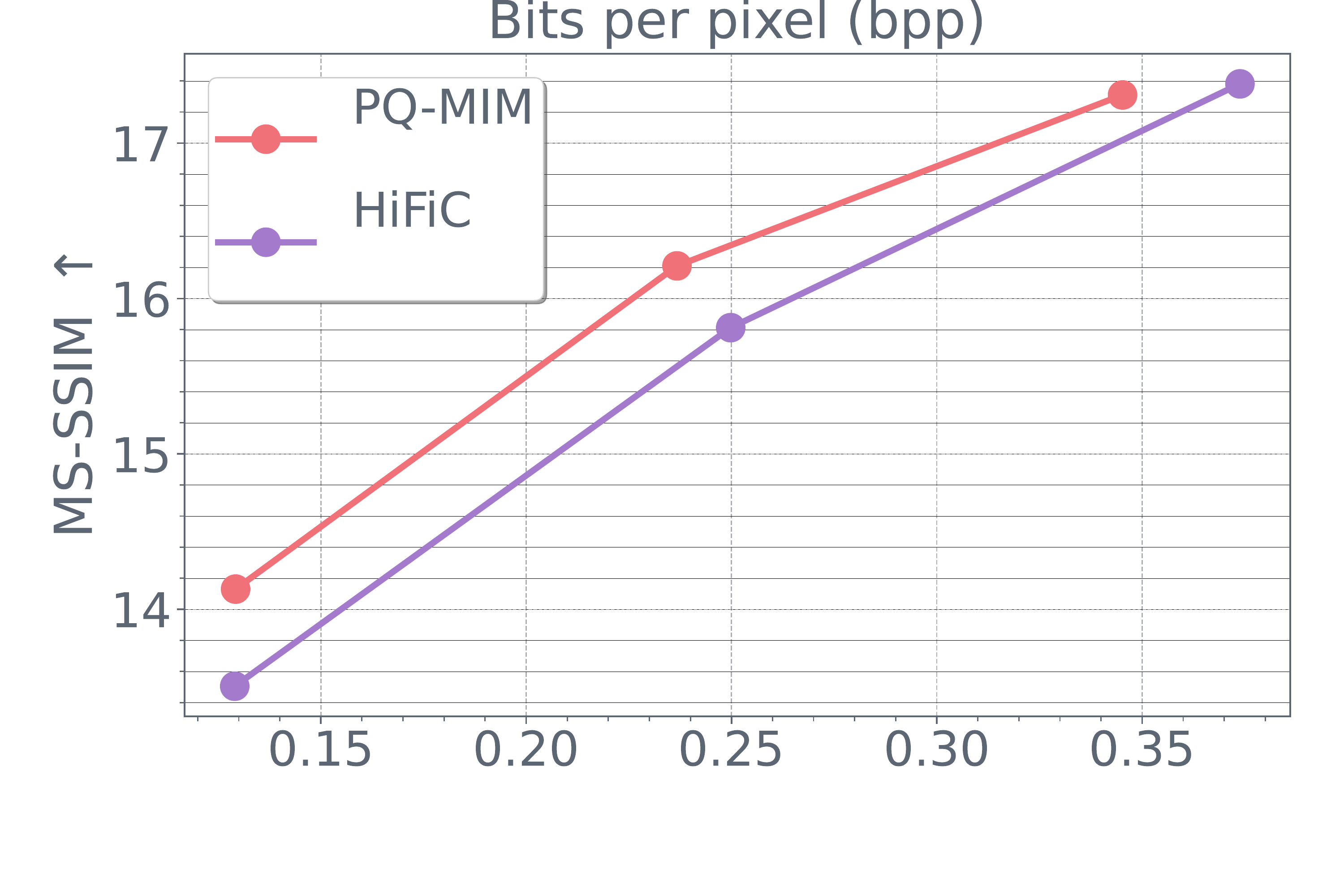}
    \end{subfigure}
    \caption{\textbf{Perceptual training and evaluation using \clic2020 test-set.} Performance of our adversarially trained \ours w.r.t percpetual metrics compared to  HiFiC \cite{mentzer20nips}. \ours provides a stronger performance on FID, KID and MS-SSIM across all reported operating points.}
    \label{fig:fid}
\end{figure*}

We first present our experimental setup in \sect{setup} and the present our results in \sect{results}. 
We provide  ablation studies in Section \ref{sec:ablations}. 
We will share our code and models.

\subsection{Experimental setup}
\label{sec:setup}

\noindent \textbf{Rate-distortion control.} For all our experiments we fix the codebook size $V=256$ and only vary the number of sub-vectors $M \in \{2, 4, 6\}$ for two different down-sampling factors $f \in \{8,16\}$.

\mypar{PQ-VAE implementation details.} 
Our PQ-VAE training uses the straight-through estimator \citep{bengio2013estimating} to propagate gradients through the quantization bottleneck. As for the quantization, the elements of the latent representation $\mathbf{z}$ are first linearly projected to a low dimensional look-up vector (dim=8 per sub-vector) followed by  $\ell_{2}$ normalization, following~\citet{yu21iclr}.
We train our model using ImageNet \citep{deng09cvpr} for 50 epochs with a batch size of 256. We use an AdamW \citep{loshchilov2017decoupled} optimizer with a peak learning rate of $1.10^{-3}$, weight decay of $0.02$ and $\beta_{2}=0.95$. We apply a linear warmup for the first 5 epochs of training followed by a cosine decay schedule for the remaining 45 epochs to a minimum learning rate of $5.10^{-5}$. Unless mentioned otherwise, for all experiments, the encoder and decoder use an  XCiT-L6 with 6 layers and hidden dimension of 768. We use sinusoidal positional embedding \citep{vaswani17nips} such that our model can flexibly operate on variable sized images. 

For the results reported in Figure~\ref{fig:results_psnr_msssim}, the models are trained using solely MSE ($\eta=0.5$) or MS-SSIM ($\eta=10.0$) distortion losses for their corresponding plots. As for the models trained with perceptual objectives (Figure~\ref{fig:fid} and Table~\ref{tab:discriminator}), they are trained with a weighted sum of MSE, LPIPS ($\alpha=1$) and adversarial loss ($\gamma=0.1$). We use a ProjectedGAN Discriminator~\citep{sauer2021projected} architecture. The perceptual training is initialized with an MSE only trained checkpoint and trained for 50 epochs using ImageNet with a learning rate of $10^{-4}$ and weight decay of $5.10^{-5}$. Similar to HiFiC~\citep{mentzer20nips}, we freeze the encoder during the perceptual training. Additionally, we find that clipping the gradient norm to a maximum value of $4.0$ improves the training stability. Our discriminator takes only the decoded image as an input and does not rely on any other conditional signal.

\begin{figure}
    \centering
    \vspace{-5mm}
    \includegraphics[width=\linewidth]{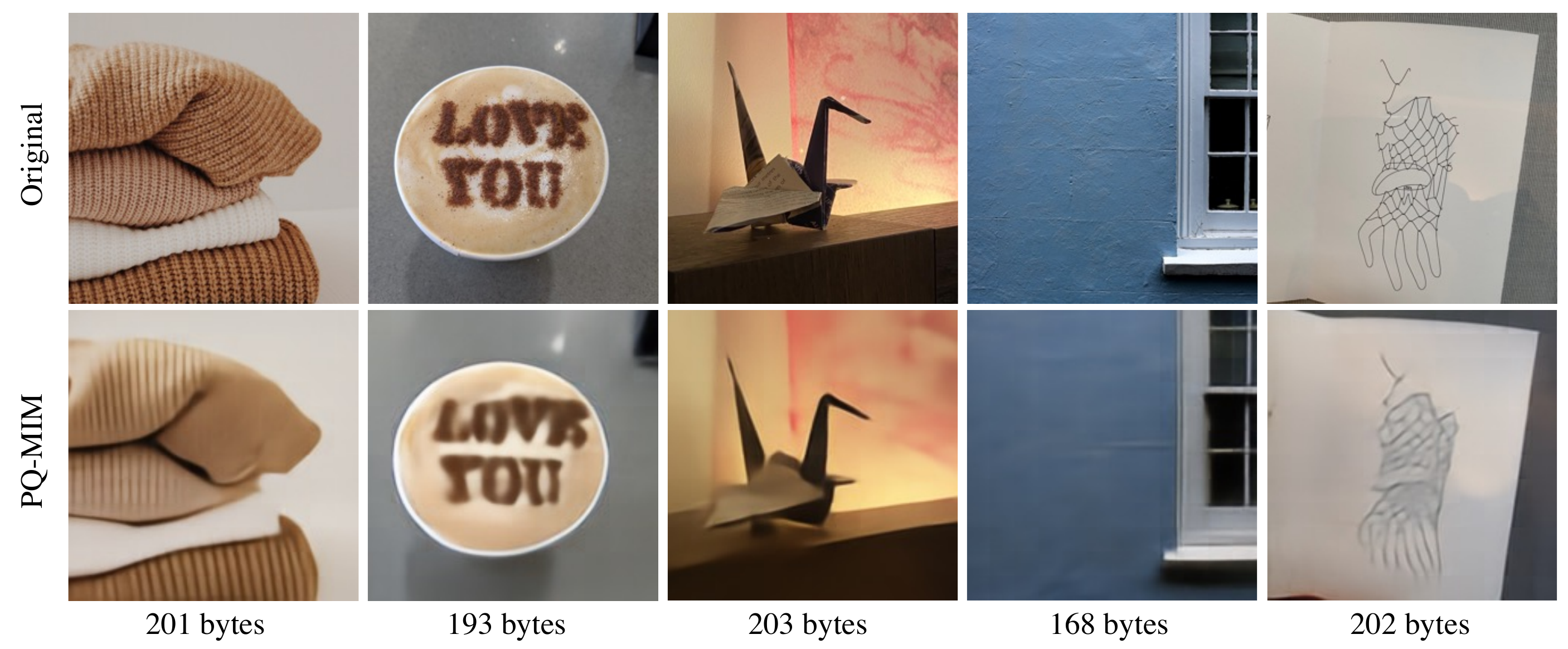}
    \caption{\textbf{Extreme Image Compression.} \ours exhibits non-trivial compression performance at the extreme compression regime (e.g. 0.03 bpp), leading to compressed image codes that can fit in a short tweet (280 characters).}
    \label{fig:extreme_compression}
\end{figure}

\mypar{MIM implementation details.} Our MIM module is an XCiT-L12 with 12 layers and embedding dimension of 768. MIM is trained simultaneously with the PQ-VAE, but the gradients are not backpropagated to the encoder or the quantizer parameters.
The PQ indicies $q$ are split into inputs and targets for the MIM model as defined by the  quincunx partitioning pattern.
By default we use $S=5$ stages.  All stages are processed with the same MIM model.
The masked patches are replaced by a learnable ``mask'' token embedding. 
The  loss is computed only for the masked patches. Since every token is assigned $M$ PQ indices, the output of the MIM transformer is passed to $M$ separate linear heads to predict $M$ softmax normalized distributions over their corresponding codebooks.
The marginal distribution of the codebook is computed as a normalized histogram over the ImageNet training set. For entropy coding, we use the implementation of the \texttt{torchac}\footnote{\scriptsize \url{https://github.com/fab-jul/torchac}} arithmetic coder.

\mypar{Datasets.} We train our models using ImageNet~\citep{deng09cvpr}. For data augmentation, we apply random resized cropping to \myres{256} images and horizontal flipping. 
For evaluation and comparison to prior work, we use \kodak~\citep{kodak1993kodak} and \tecnick~\citep{tecnick} datasets for PSNR and MS-SSIM. Moreover, we compute the perceptual metrics (FID~\citep{heusel17nips}, KID~\citep{binkowski2018demystifying}) for perceptually trained models using the \clic2020 test-set~\citep{CLIC2020} (428 images) using the same patch cropping scheme detailed by \cite{mentzer20nips}.

\mypar{Baselines.}
We compare to several existing neural compression baselines: the scale hyperprior model~\citep{balle18iclr}, mean \& scale hyperprior~\citep{minnen18nips}, GMM hyperprior~\citep{cheng20cvpr}, and HiFiC~\citep{mentzer20nips}.
Among the non-neural codecs, we compare to the popular BPG \citep{ballard_bpg}, WebP, and JPEG2000.

\subsection{Main experimental results}
\label{sec:results}

\noindent \textbf{Comparison to existing (neural) codecs.}
We compare \ours to other approaches across a wide range of bitrates in Figure~\ref{fig:results_psnr_msssim}\footnote{All results for the baselines are reported using the authors official repositories}.
Note that in our evaluation we consider bitrates that are  an order of magnitude lower than what is typically studied in the literature: most previous studies were limited to 0.1 bpp and above, see \eg~\citep{balle18iclr,cheng20cvpr,minnen18nips}.
The extremely low bitrates we consider make it possible to transmit a  \myres{256} image in an SMS or a tweet (280 characters)\footnote{For example, a bitrate of 0.03 yields $256\times 256 \times 0.03 / 8\approx 246$ bytes for a \myres{256} image.} as shown in Figure ~\ref{fig:extreme_compression}. \ours achieves a strong and competitive performance for both \kodak and \tecnick datasets, outperforming all prior neural and standard codecs with the exception of GMM hyperprior~\citep{cheng20cvpr}. We observe that \ours is particularly strong for low rates, making it a good fit for extreme compression scenarios. Moreover, \ours exhibits a particularly strong MS-SSIM performance which was designed to model the human visual contrast perception \citep{wang2004image, wang2003multiscale}.

\mypar{Perceptual metrics comparison.} In Figure~\ref{fig:fid},  we compare \ours to HiFiC~\citep{mentzer20nips} in perceptual quality measures like FID and KID as well as MS-SSIM.
HiFiC is based on the mean \& scale hyperprior model~\citep{minnen18nips}, but adds adversarially trained discriminator model to improve the perceptual quality of the image reconstructions. \ours, with perceptual training, achieves a strong performance for all reported metrics, outperforming HiFiC for all operating points.

\begin{table*}[t!]
 \smallskip
    \begin{minipage}[b]{0.48\linewidth}
         \centering
        \caption{\textbf{Discriminator Architecture}. 
        We investigate multiple discriminator including StyleGAN \citep{stylegan}, ProjectedGAN \citep{sauer2021projected} and UNet \citep{schonfeld2020u}. \label{tab:heads}}
        \scalebox{0.85}{
        \begin{tabular}{l c c c }
            \toprule
              Discriminator & FID $\downarrow$ & KID $\downarrow$ & MS-SSIM $\uparrow$ \\
              \midrule
              None & 26.1 & 1.2$\times{-2}$ & 15.3 \\
             \midrule
              StyleGAN & \textbf{3.57} & \textbf{4.8$\times10{-4}$} & 13.3\\
              ProjectedGAN  & 3.69  & 6.6$\times10{-4}$ & \textbf{14.1} \\
              UNet & 3.87 & 6.7$\times10{-4}$ &  13.9 \\
          \bottomrule
          \label{tab:discriminator}
        \end{tabular}}
        
    \end{minipage}%
    \hfill
    \begin{minipage}[b]{0.5\linewidth}
        \centering
        \caption{\textbf{Different MIM masking policies.} \ours with quincunx pattern with 5 steps reduces the bpp significantly (27\%). Additionally, \ours is orders of magnitude cheaper in terms of FLOPs compared to an autoregressive raster order masking pattern. \label{tab:masking_policy}}
        \scalebox{0.9}{
        \begin{tabular}{l c c c}
            \toprule
             Masking policy & \#steps & bpp & MACs/Pixel (M)\\
             \midrule
             Marginal baseline & 1 & 0.512 & 0.69  \\
             Raster order & $T$ & \texttt{OOM} & 24.4$\times10^3$  \\
             Quincunx & 5 & 0.373 & 6.66  \\
             \bottomrule
        \end{tabular}}

    \end{minipage}%
\end{table*}

\subsection{Analysis and Ablations}
\label{sec:ablations}

\begin{figure*}[t!]
        \vspace{-2mm}
        \begin{minipage}[t]{0.48\linewidth}
        \centering
            \includegraphics[width=1.0\linewidth]{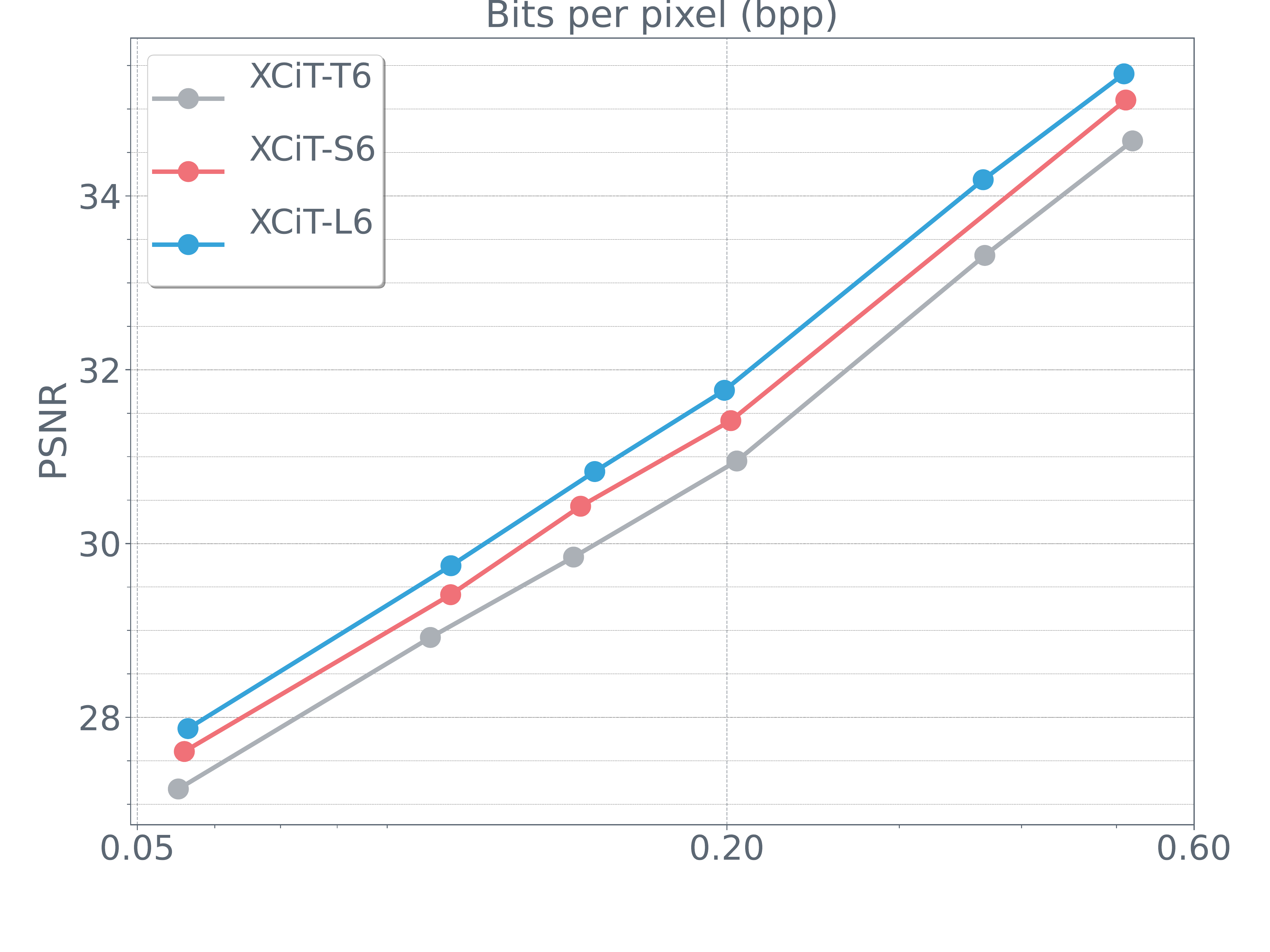}
        \caption{\textbf{Autoencoder capacity.} The RD performance of different encoder/decoder capacities. We use the same model trunk for (i) the encoder; (ii) the PQ-MIM entropy model ; and (iii) the decoder. Increasing the model size from a XCiT-T6 model (3.5M params) to XCiT-L6 (47M params) increases the performance by typically +0.8dB.} %
        \label{fig:autoencoder_capacity}
    \end{minipage}
    \hfill
    \begin{minipage}[t]{0.5\linewidth}
            \centering
            \includegraphics[width=1.0\linewidth]{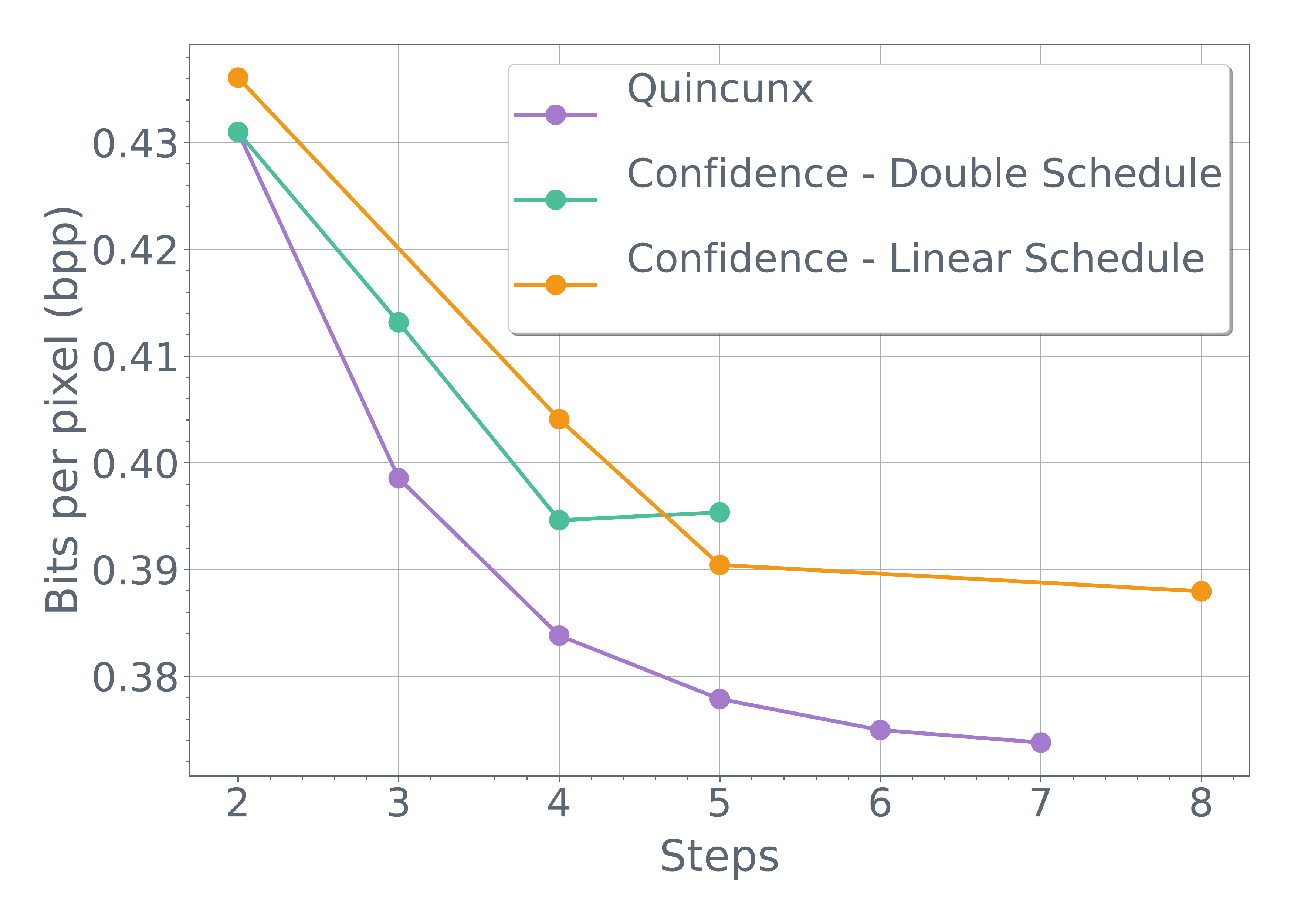}
            \caption{\textbf{MIM number of steps.} In addition to quincunx, we explore using the prediction confidence as the patch masking policy following MaskGiT~\citep{chang22arxiv}. We test a linear and doubling schedules. Quincunx provides a higher rate saving, even compared to confidence policy with a longer schedule.} 
            \label{fig:mim_steps}
    \end{minipage}
\end{figure*}

\noindent \textbf{Model size and architecture.}
In Figure~\ref{fig:autoencoder_capacity}, we analyze the effect of using XCiT of different capacities for the autoencoder with respect to rate-distortion trade-off. We observe that the performance improves with higher capacity autoencoders, but their is a diminishing return with further increase in capacity. For all our experiments we use an XCiT-L6 since it achieves the best performance.

\mypar{Masking patterns.}
In Table~\ref{tab:masking_policy}, we compare to predicting tokens one-by-one autoregressively in a raster-scan order, the same pattern used in VQ-VAE based generative image models such as DALL-E~\citep{ramesh21arxiv} and VQ-GAN~\citep{esser21cvpr}. In contrast to \ours, raster-scan 
models require causal attention, which makes XCiT not a good fit. We use a standard ViT model instead. However, due to the quadratic complexity of ViT and the high resolution of images typically used for evaluation of compression method (e.g. \tecnick), our autoregressive variant consistently exceeded the memory limits, even when using \texttt{A100} GPUs with 40GB memory. Moreover, raster scan fully-autoregressive models  results in extremely  expensive FLOP count since it needs $T$ separate forward passes per image.
On the other hand, our stage-wise MIM with quincunx pattern requires  4 evaluations, and does not scale with the  image resolution as is the case for raster, making it a more practical solution.

\mypar{Number of prediction stages.}
We compare the quincunx masking pattern with masking based on confidence score following MaskGiT~\citep{chang22arxiv} patch selection procedure for image generation. 
In the latter case, at a given step, we pick the patches to transmit dynamically based on their confidence score. 
The confidence is  defined as the maximum across the probabilities assigned over the vocabulary by the model.\footnote{  
Note that for decoding the same confidence score can be used to identify the group of tokens to decode.} 
For the quincunx and confidence based masking policy we use the same 5-stage scheme, in which the number of tokens in subsequent groups doubles in size. We ablate the number of prediction stages $S$  for the quincunx and confidence-based sampling. 
For the latter we consider two options: (i) a linear scheme where each group of tokens contains (approximately) the same number of tokens; and (ii) a doubling scheme where each subsequent group of tokens is double the size of the previous group, as is also used for  the quincunx pattern.
Every point on the curves in  \fig{mim_steps}  corresponds to the bitrate when  encoding/decoding with a given number of steps. 
For example, $S=2$ steps means we have only two steps each encoding/decoding 50\% of the patches. For the 3-steps  doubling schedule the groups have sizes  of 1/4, 1/4, 1/2, and so on. 
For all three patterns the bitrate  monotonically decreases with the number of steps, as more  tokens can be predicted from larger contexts.

On the one hand, using confidence-based masking patterns, doubling and linear schemes lead to mostly comparable bitrates for the same number of steps, with further improvement for linear with more steps however with diminishing return. On the other hand, quincunx provides a stronger reduction ratio with strictly 5 steps, even when compared with confidence-based masking with higher number of steps.

\begin{figure*}
    \setlength{\tabcolsep}{1pt}
    \centering
    \vspace{-5mm}
    \begin{tabular}{ccccc}
    \texttt{No Patch Dropping} & \texttt{10\% Drop Rate} & \texttt{20\% Drop Rate} & \texttt{30\% Drop Rate} & \texttt{50\% Drop Rate} \\
     \includegraphics[width=0.19\linewidth]{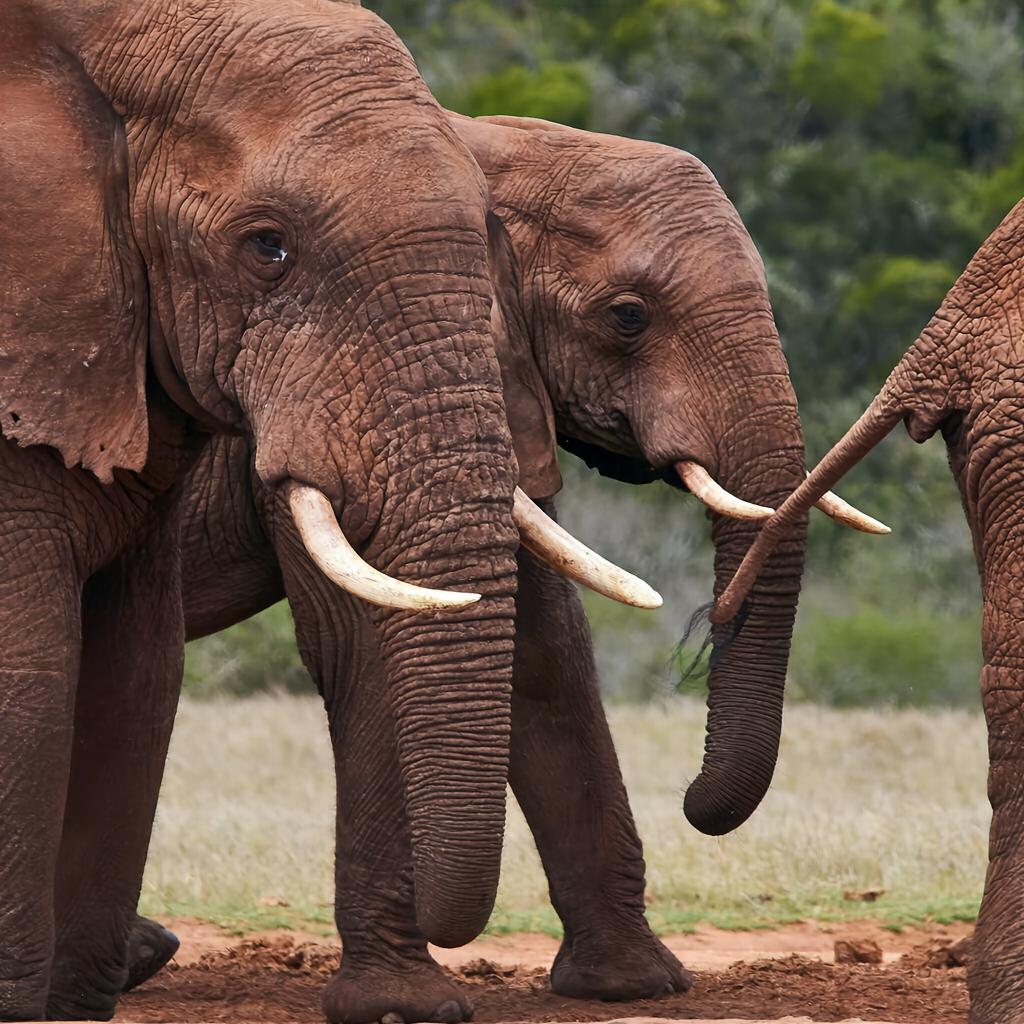} & \includegraphics[width=0.19\linewidth]{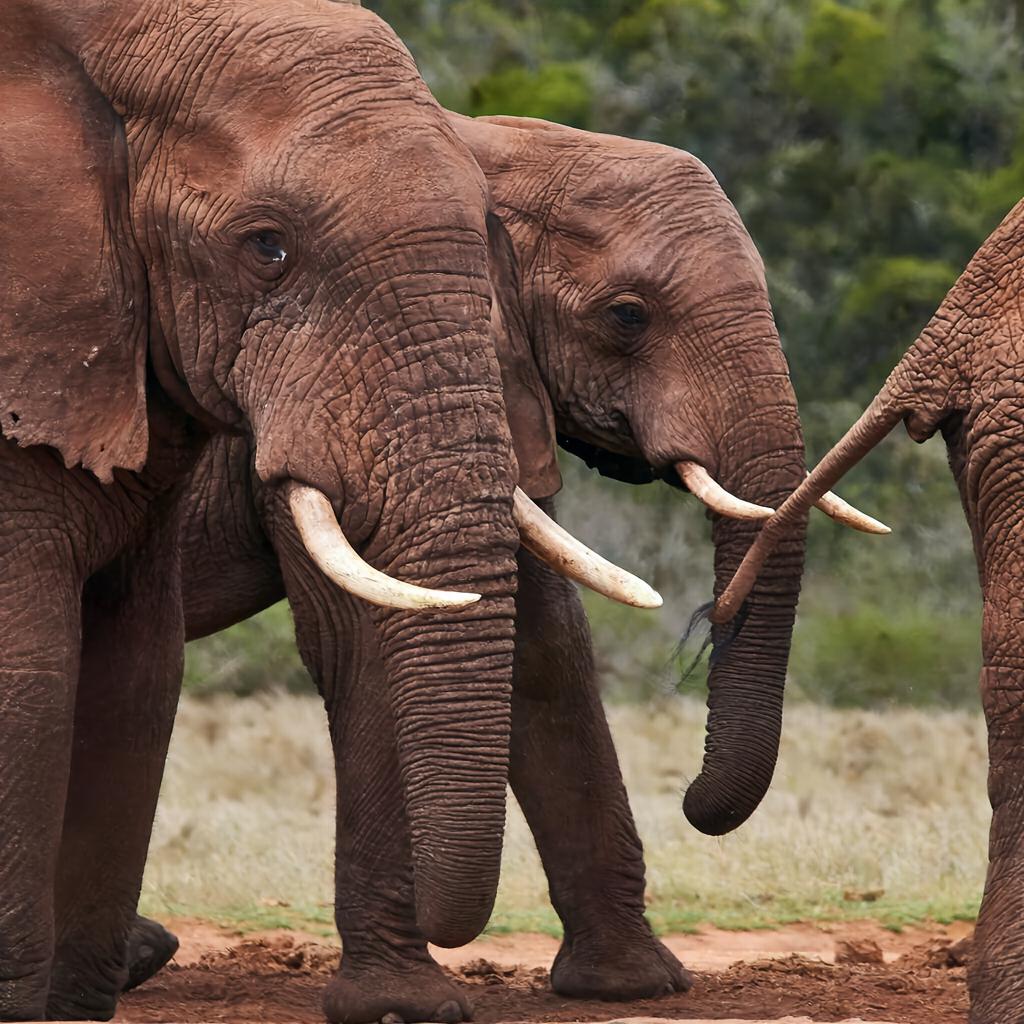} & \includegraphics[width=0.19\linewidth]{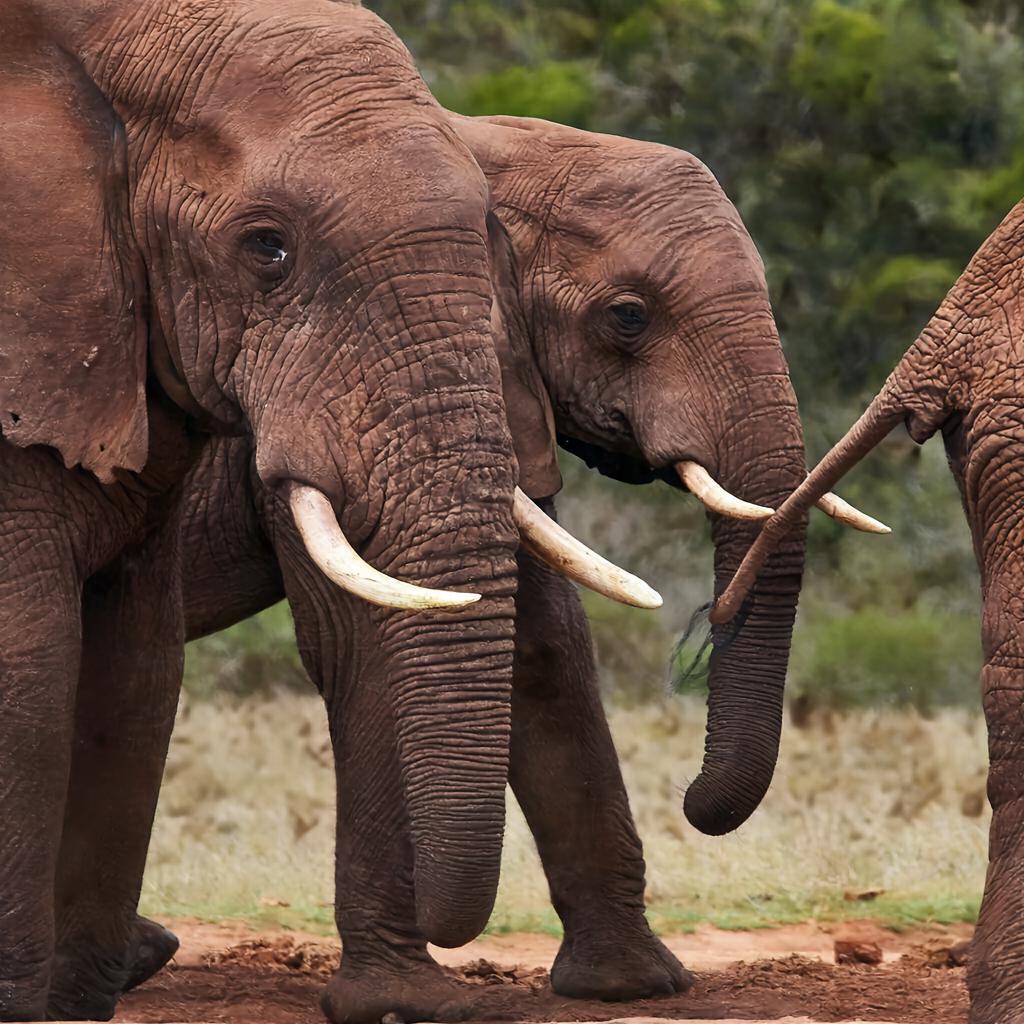} & \includegraphics[width=0.19\linewidth]{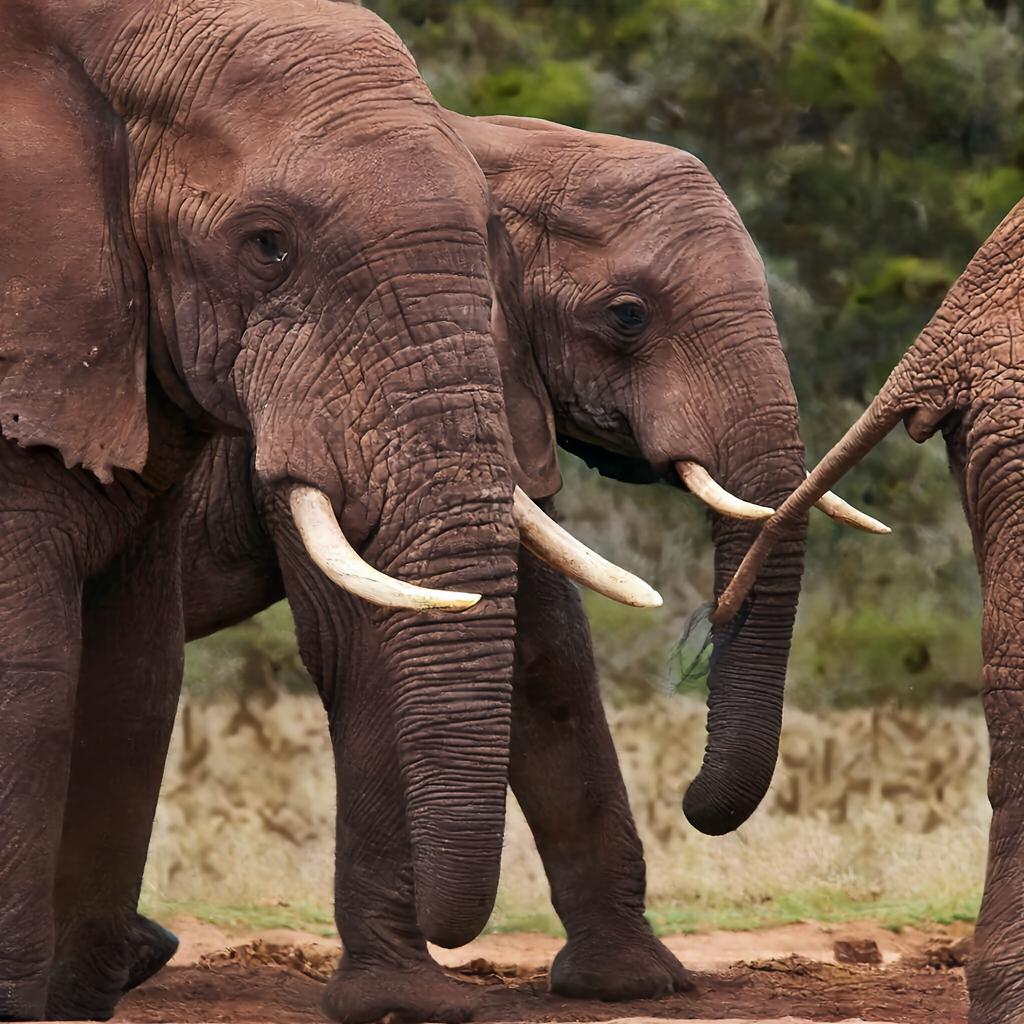} & \includegraphics[width=0.19\linewidth]{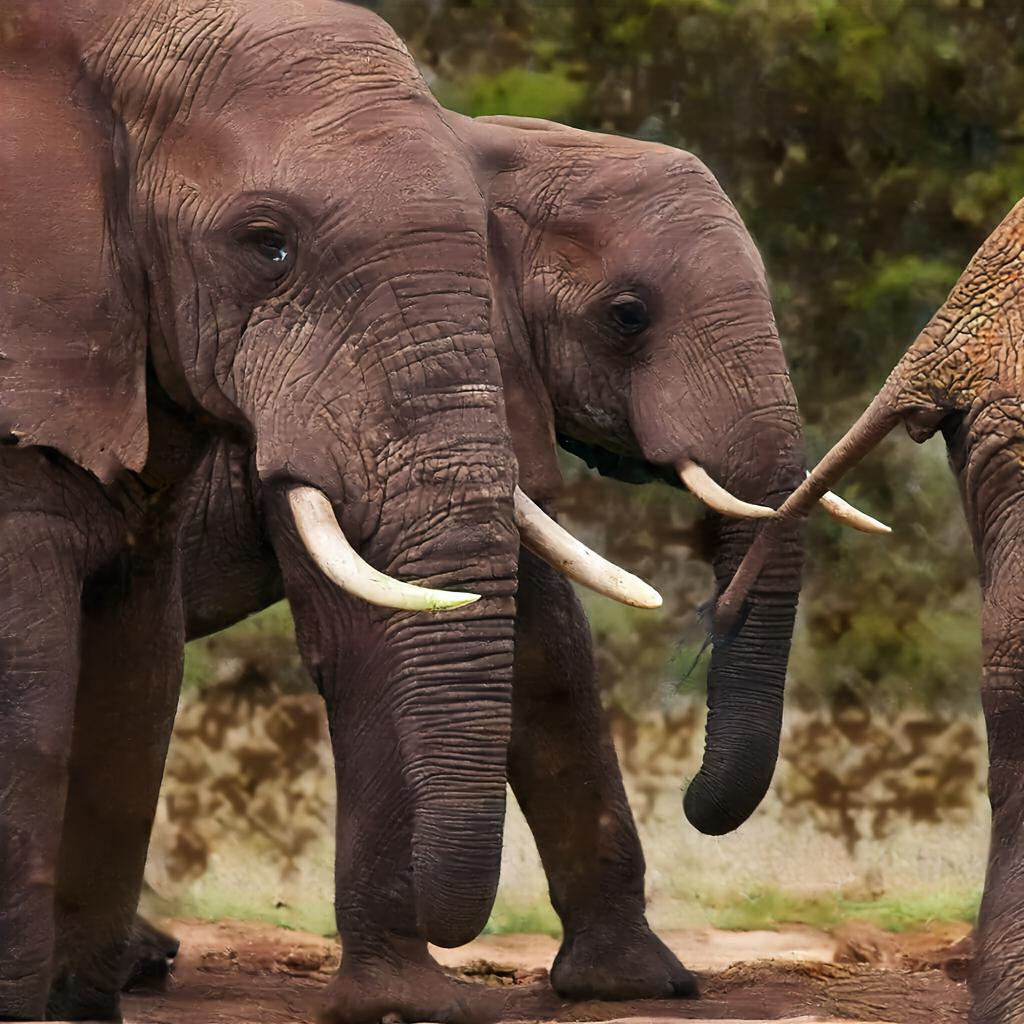} \\
      \includegraphics[width=0.19\linewidth]{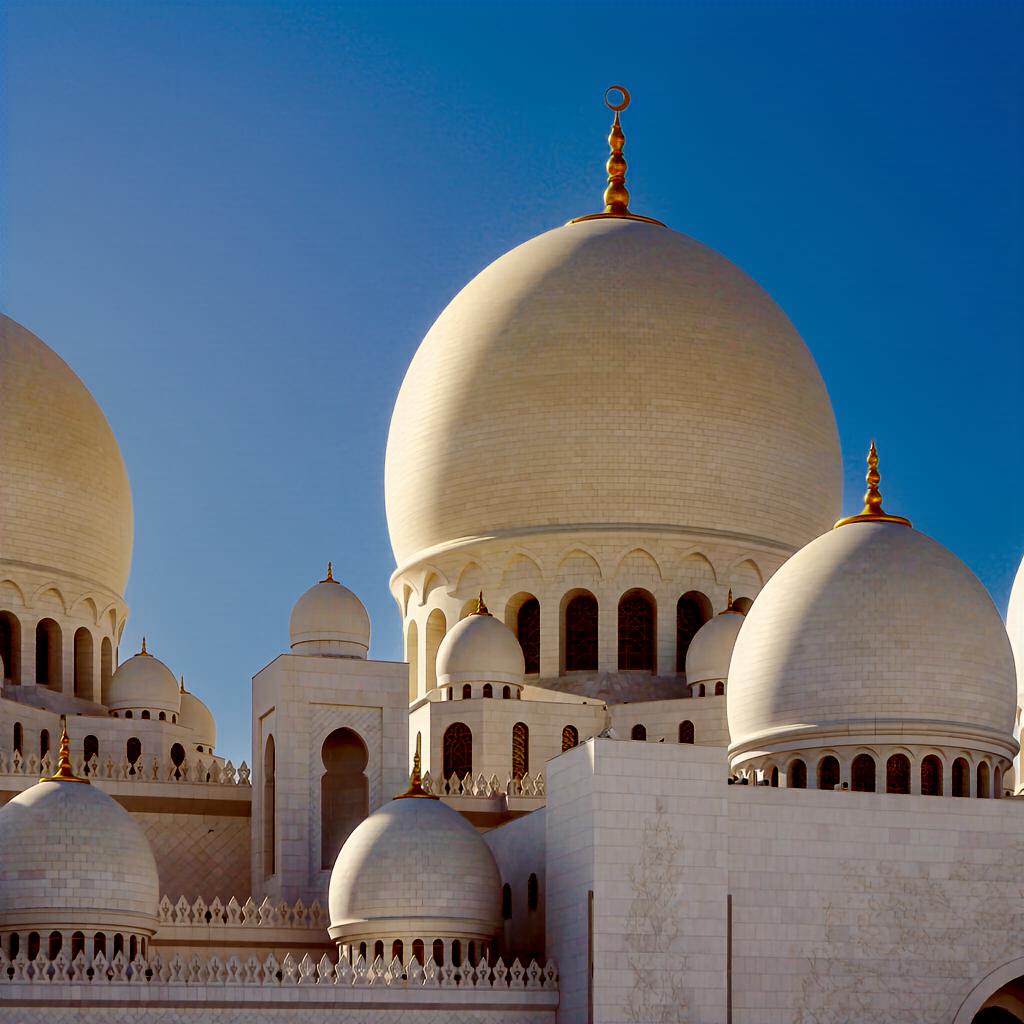} & \includegraphics[width=0.19\linewidth]{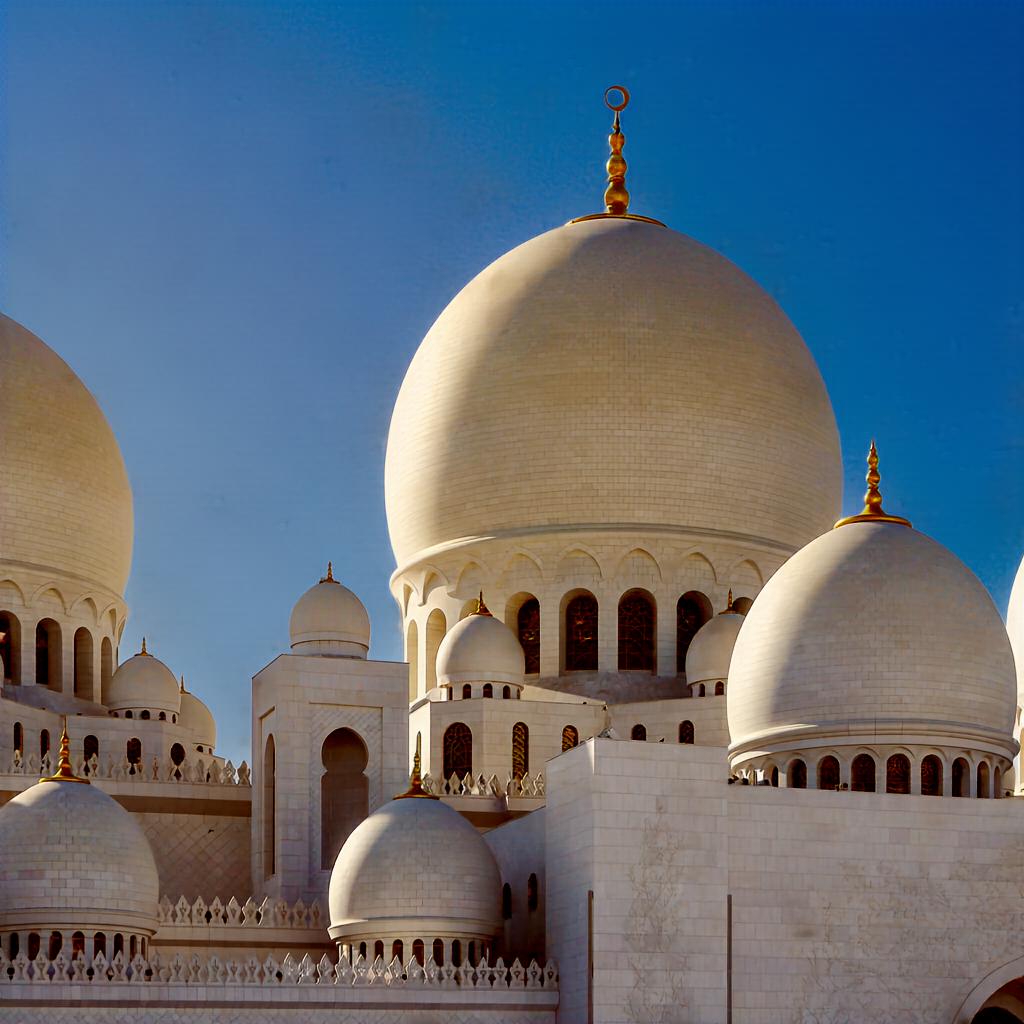} & \includegraphics[width=0.19\linewidth]{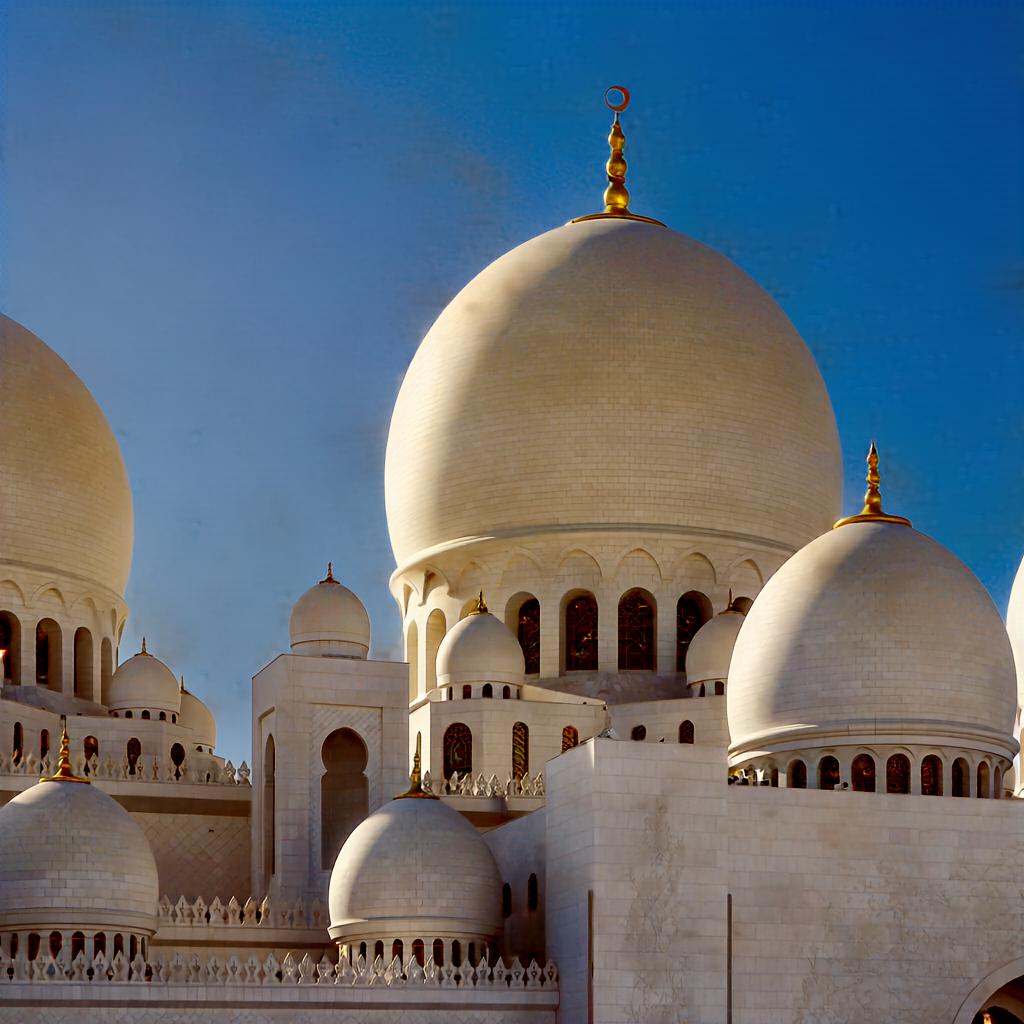} & \includegraphics[width=0.19\linewidth]{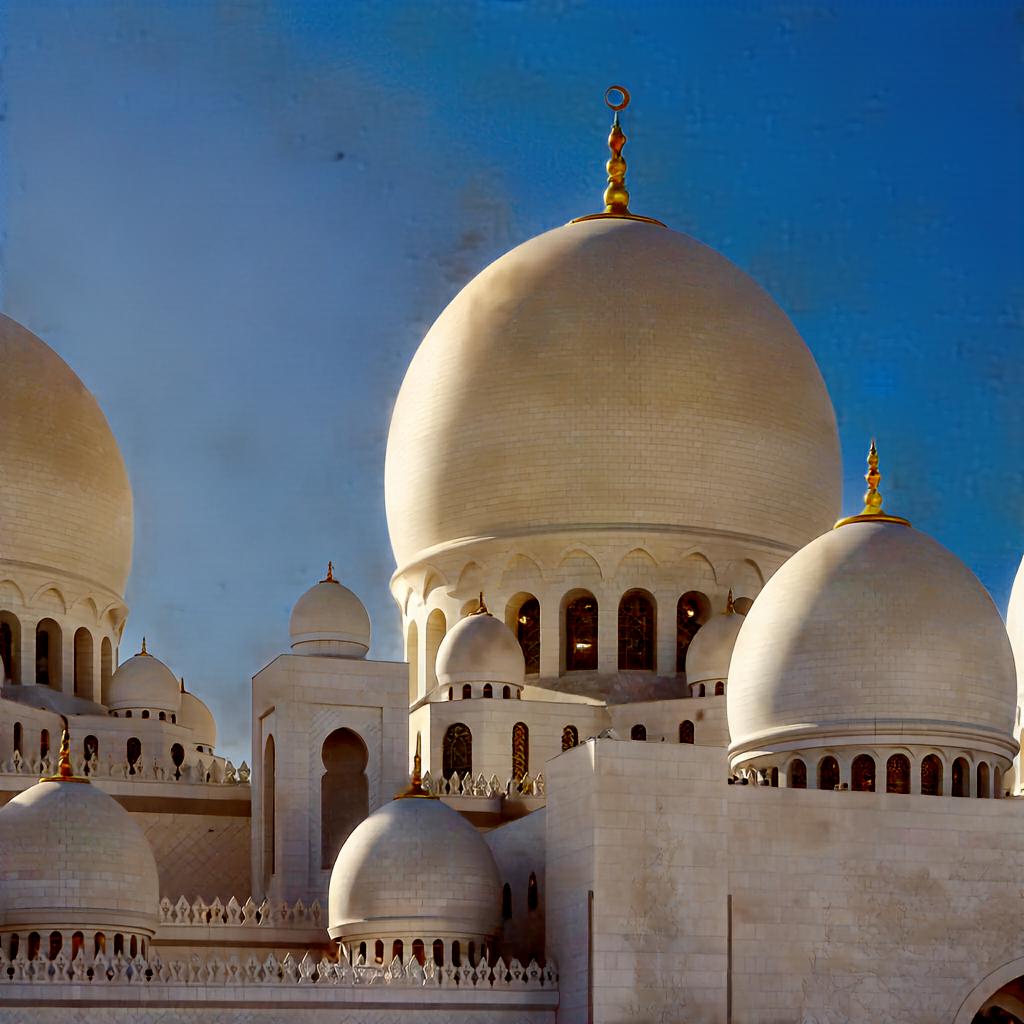} & \includegraphics[width=0.19\linewidth]{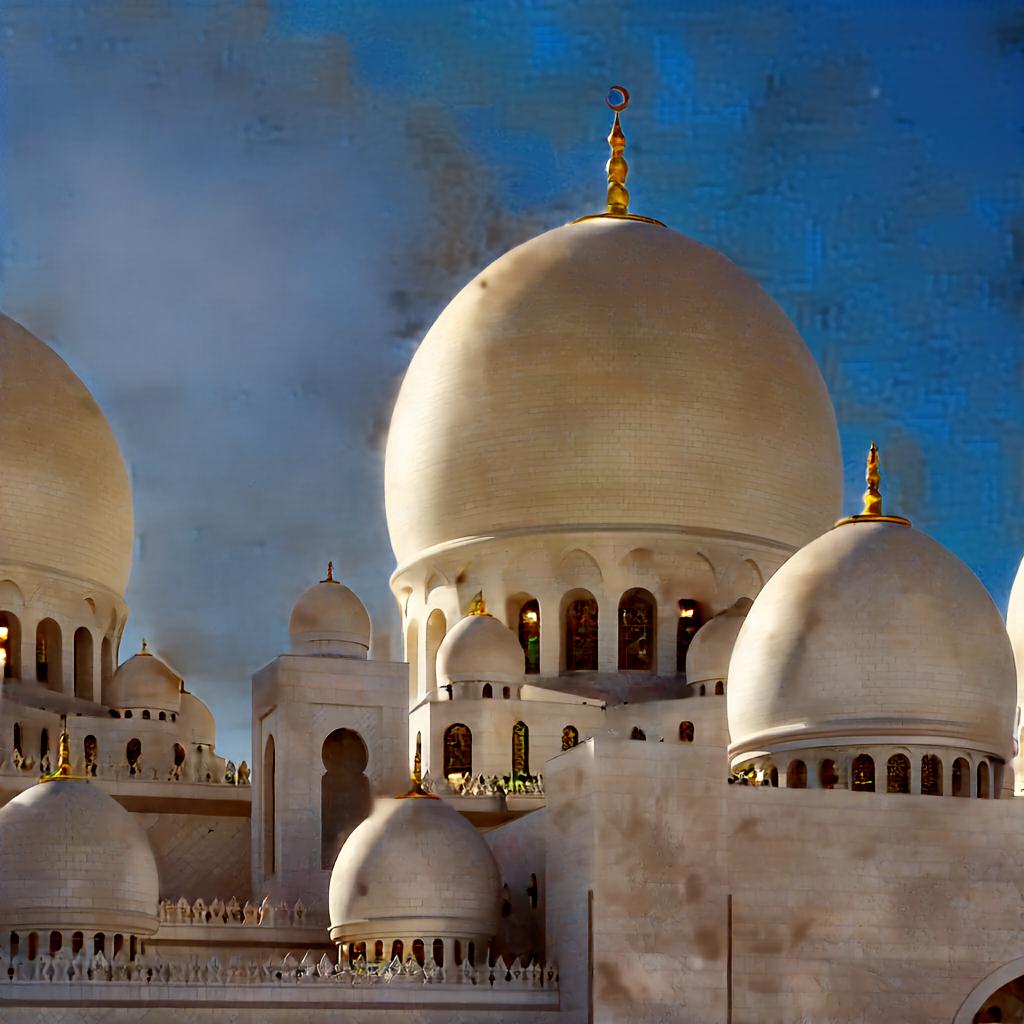} \\
    \end{tabular}
    \caption{\textbf{\ours can operate on a partial set of transmitted tokens.} Since \ours is compatible with generation, the conditional entropy model can be repurposed to predict the PQ codes for missing parts in an image. We show results for different dropping rates of transmitted tokens. \ours exhibits strong inpainting abilities. Even for the extreme case where half of the image patches are dropped, \ours can still retain a large percentage of the original image structure and details.
    \label{fig:inpaiting_rand_bitstream}}
   \vspace{-2mm}
\end{figure*}

\mypar{Predicting missing patches.} 
When encoding/decoding the discrete image representation $\mathbf{q}$, we reduce the bitrate with our models by their ability to predict the remaining tokens given those of preceding stages.
To illustrate the predictive ability of our model, we consider an experiment where we remove a subset of the tokens (sampled randomly), and use our model to fill in the missing patches by conditioning on the observed ones.
We then use the PQ-VAE decoder to decode the discrete latent codes including the filled in ones. 
In \fig{inpaiting_rand_bitstream} we show the results obtained when removing 10\%, 20\%, 30\% and 50\% of the patches.
Our MIM is able to model redundancies among the patches, which corroborates our findings of its compression ability.

\subsection{Limitations}

Image compression, and compression of visual data in general, is an important technology to scale the distribution of visual data. This is ever more  important to cope with the growing quantity of visual data that is streamed in the form of video and for augmented and virtual reality applications. 
Compression is also critical to allow users with low-bandwidth connections to benefit from applications relying on sharing of image, video, or virtual reality data.

\mypar{Caveats of learned neural compression models: biases and performance. }
As with any machine learning model, potential biases in the training data may be transferred into the model via training. 
In our setting, this can affect the autoencoding reconstruction abilities for content under-represented in the training data, as well as the compression abilities of the model for such data. Such biases should be assessed before deployment of the model.
Beyond rate-distortion trade-offs, important evaluation dimensions include the energy and latency performance of compression models. 
Current neural compression methods, including ours, need to be further optimized to be competitive with existing codecs on these criteria. 

\mypar{Specific limitations of our approach. } 
In our work, we specifically focus on a high compression regime, with compression rate lower than 0.6 bit per pixel. This is a favorable case for our approach: it benefits from the generative model capability inherited from the VQ-VAE. Compared to VQ, PQ allows for higher rates, but it becomes comparatively less effective when increasing the number of subquantizers, as one can deduce from the slopes of the rate-distortion curves in Fig.~\ref{fig:results_psnr_msssim}. It is also computationally more expensive.

\section{Conclusion}
\label{sec:conc}
In this paper, we have revisited vector quantization for neural image compression. We introduced a product-quantization variants of VQ-VAE and shown that it has a better scaling properties in terms of bit-rate. Additionally, we introduced a novel conditional entropy model based on masked image modeling. We have shown that combined with the qunicunx partitioning pattern, \ours provides strong reduction in bit-rate. \ours exhibits a competitive performance for PSNR and MS-SSIM metric compared to strong neural compression baselines. Furthermore, when we train \ours using perceptual losses, it provides a strong performance on multiple metrics (e.g. FID, KID and MS-SSIM) compared the strong baseline of HiFiC. Finally, we have shown that \ours can operate in a hybrid compression/generation mode where it can fill the gaps for non-transmitted patches.

\begingroup
    \setlength{\bibsep}{4pt}
    \bibliography{refs}
    \bibliographystyle{tmlr}
\endgroup

\clearpage

\appendix
                                                                                                                                                                                                                                                                                                                                                                                                                                                                                                                                                                                                                                                                                                                                                                                                                                                                                                                                                                                                                                                                                                                                                                                                                                                                                                                                                                                                                                                                                                                                                                                                                                                                                                                                                                                                                                                                                                                                                                                                                                                                                                                                                                                                                                                                                                                                                                                                                                                                                                                                                                                                                                                                                                                                                                                                                                                                                                                                                                                                                                                                                                                                                                                                                                                                                                                                                                                                                                                                                                                                                                                                                                                                                                                                                                                                                                                                                                                                                                                                                                                                                                                                                                                                                                                                                                                                                                                                                                                                                                                                                                                                                                                                                                                                                                                                                                                                                                                                                                                                                                                                                                                                                                                                                                                                                                                                                                                                                                                                                                                                                                                                                                                                                                                                                                                                                                                                                                                                                                                                                                                                                                                                                                                                                                                                                                                                                                                                                                                                                                                                                                                                                                                                                                                                                                                                                                                                                                                                                                                                                                                                                                                                                                                                                                                                                                                                                                                                                                                                                                                                                                                                                                                                                                                                                                                                                                                                                                                                                                                                                                                                                                                                                                                                                                                                                                                                                                                                                                                                                                                                                                                                                                                                                                                                                                                                                                                                                                                                                                                                                                                                                                                                                                                                                                                                                                                                                                                                                                                                                                                                                                                                                                                                                                                                                                                                                                                                                                                                                                                                                                                                                                                                                                                                                                                                                                                                                                                                                                                                                                                                                                                                                                                                                                                                                                                                                                                                                                                                                                                                                                                                                                                                                                                                                                                                                                                                                                                                                                                                                                                                                                                                                                                                                                                                                                                                                                                                                                                                                                                                                                                                                                                                                                                                                                                                                                                                                                                                                                                                                                                                                                                                                                                                                                                                                                                                                                                                                                                                                                                                                                                                                                                                                                                                                                                                                                                                                                                                                                                                                                                                                                                                                                                                                                                                                                                                                                                                                                                                                                                                                                                                                                                                                                                                                                                                                                                                                                                                                                                                                                                                                                                                                                                                                                                                                                                                                                                                                                                                                                                                                                                                                                                                                                                                                                                                                                                                                                                                                                                                                                                                                                                                                                                                                                                                                                                                                                                                                                                                                                                                                                                                                                                                                                                                                                                                                                                                                                                                                                                                                                                                                                                                                                                                                                                                                                                                                                                                                                                                                                                                                                                                                                                                                                                                                                                                                                                                                                                                                                                                                                                                                                                                                                                                                                                                                                                                                                                                                                                                                                                                                                                                                                                                                                                                                                                                                                                                                                                                                                                                                                                                                                                                                                                                                                                                                                                                                                                                                                                                                                                                                                                                                                                                                                                                                                                                                                                                                                                                                                                                                                                                                                                                                                                                                                                                                                                                                                                                                                                                                                                                                                                                                                                                                                                                                                                                                                                                                                                                                                                                                                                                                                                                                                                                                                                                                                                                                                                                                                                                                                                                                                                                                                                                                                                                                                                                                                                                                                                                                                                                                                                                                                                                                                                                                                                                                                                                                                                                                                                                                                                                                                                                                                                                                                                                                                                                                                                                                                                                                                                                                                                                                                                                                                                                                                                                                                                                                                                                                                                                                                                                                                                                                                                                                                                                                                                                                                                                                                                                                                                                                                                                                                                                                                                                                                                                                                                                                                                                                                                                                                                                                                                                                                                                                                                                                                                                                                                                                                                                                                                                                                                                                                                                                                                                                                                                                                                                                                                                                                                                                                                                                                                                                                                                                                                                                                                                                                                                                                                                                                                                                                                                                                                                                                                                                                                                                                                                                                                                                                                                                                                                                                                                                                                    
\normalsize
\begin{center}
\Large{\textbf{Image Compression with Product Quantized Masked Image Modeling: Supplementary Material}}
\end{center}

\appendix

\section{Lossless Compression}
\label{app:lossless}

The goal of lossless compression is to encode samples from a discrete probably distribution $x\sim p_d(x)$ $x\in \mathcal{X}$ as bit strings $m=\texttt{enc}(x) \in \{0,1\}^*$ of shortest possible length $\ell(m)$ such that $x$ can be decoded without loss of information $x = \texttt{dec}(m)$. 

Block codes assign a unique binary code of equal length to each event in $\mathcal{X}$. The expected length per symbol of such codes is $l(m)=\lceil \log_2 |\mathcal{X}| \rceil$ bits. 

We can improve this result if we consider to give shorter code words to more frequently occurring symbols and longer codes to less frequently occurring ones. In fact building on this idea, it can be shown that the smallest expected achievable average code length per point using an approximation to the true data distribution $p(x)$, is given by the \emph{cross-entropy}:
\begin{align}
    H(p_d, p) := \mathbb{E}_{x \sim p_d} \left[ -\log p(x)\right].
\end{align}
This quantity is bounded from below (seen via Jensen's inequality) by the \emph{entropy} of the ``true'' probability distribution $p_d$, i.e.\ when $p=p_d$ \citep{mackay2003information, cover2012elements}. 

Entropy coders reach this bound up to a small constant $\epsilon$. An entropy codec is a tuple of an inverse pair of functions, $\texttt{enc}_{p_{X}}$ and $\texttt{dec}_{p_{X}}$ that achieve near optimal compression rates on sequences of symbols.
\begin{equation}
\begin{aligned}
    &\texttt{enc}_{p}:m, x\mapsto m'\\
    &\texttt{dec}_{p}:m'\mapsto (m, x),
\end{aligned}
\end{equation}
such that $\ell(m')=\ell(m)+\log_2 p(x) + \epsilon $.
Specifically in this paper, we  rely on \emph{Arithmetic coding}, a type of stream coder that represents data by means of a subinterval of the unit interval $[0,1]$ of the real line.  

\clearpage
\section{Qualitative Samples}
\label{app:qual_samples}

\begin{figure}[h!]
    \begin{subfigure}[t]{0.5\linewidth}
    \centering
    \includegraphics[width=\linewidth, height=0.35\paperheight]{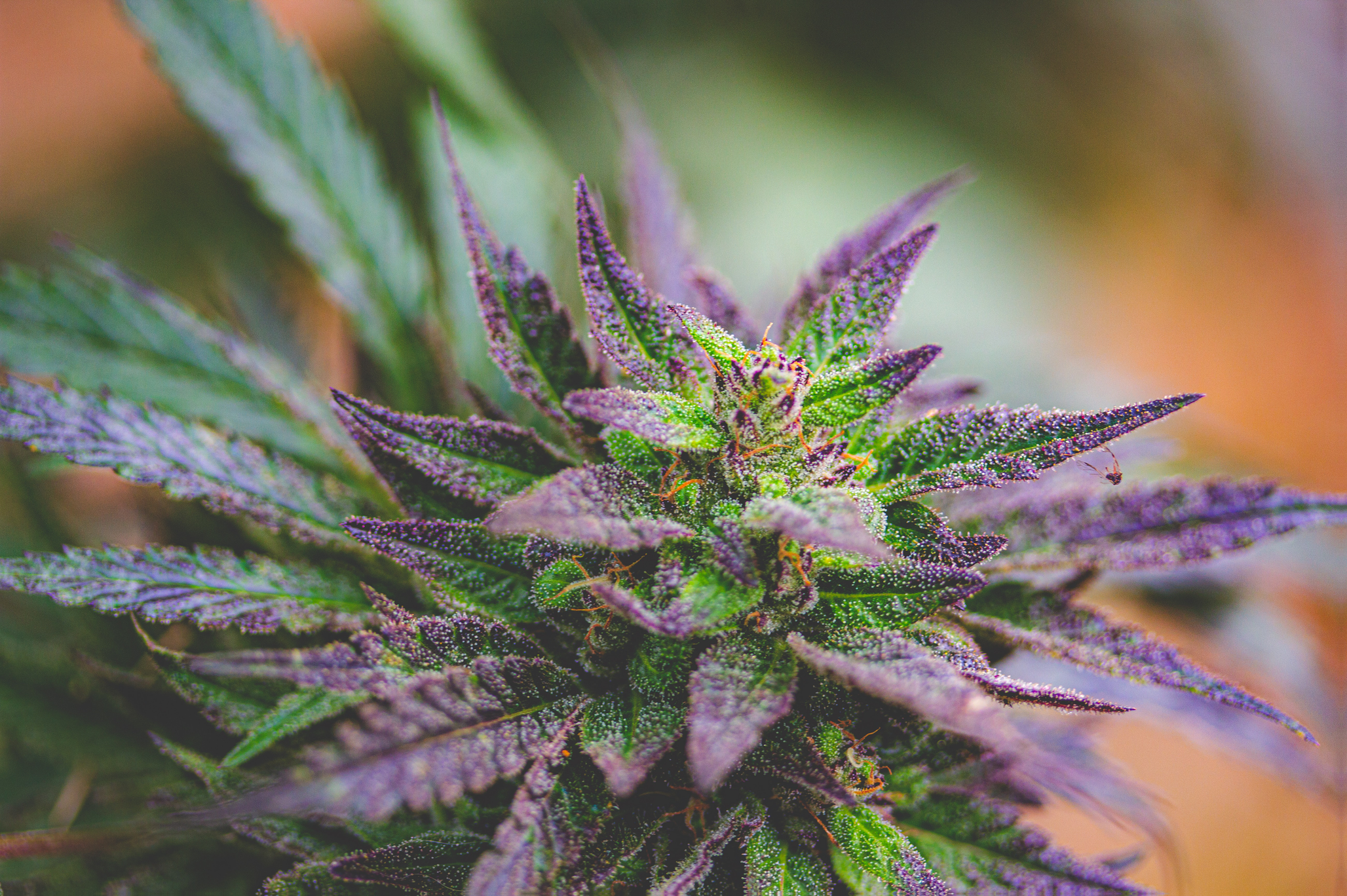}
    \caption{Original}
    \end{subfigure}
    \hfill
    \begin{subfigure}[t]{0.5\linewidth}
    \centering
    \includegraphics[width=\linewidth, height=0.35\paperheight]{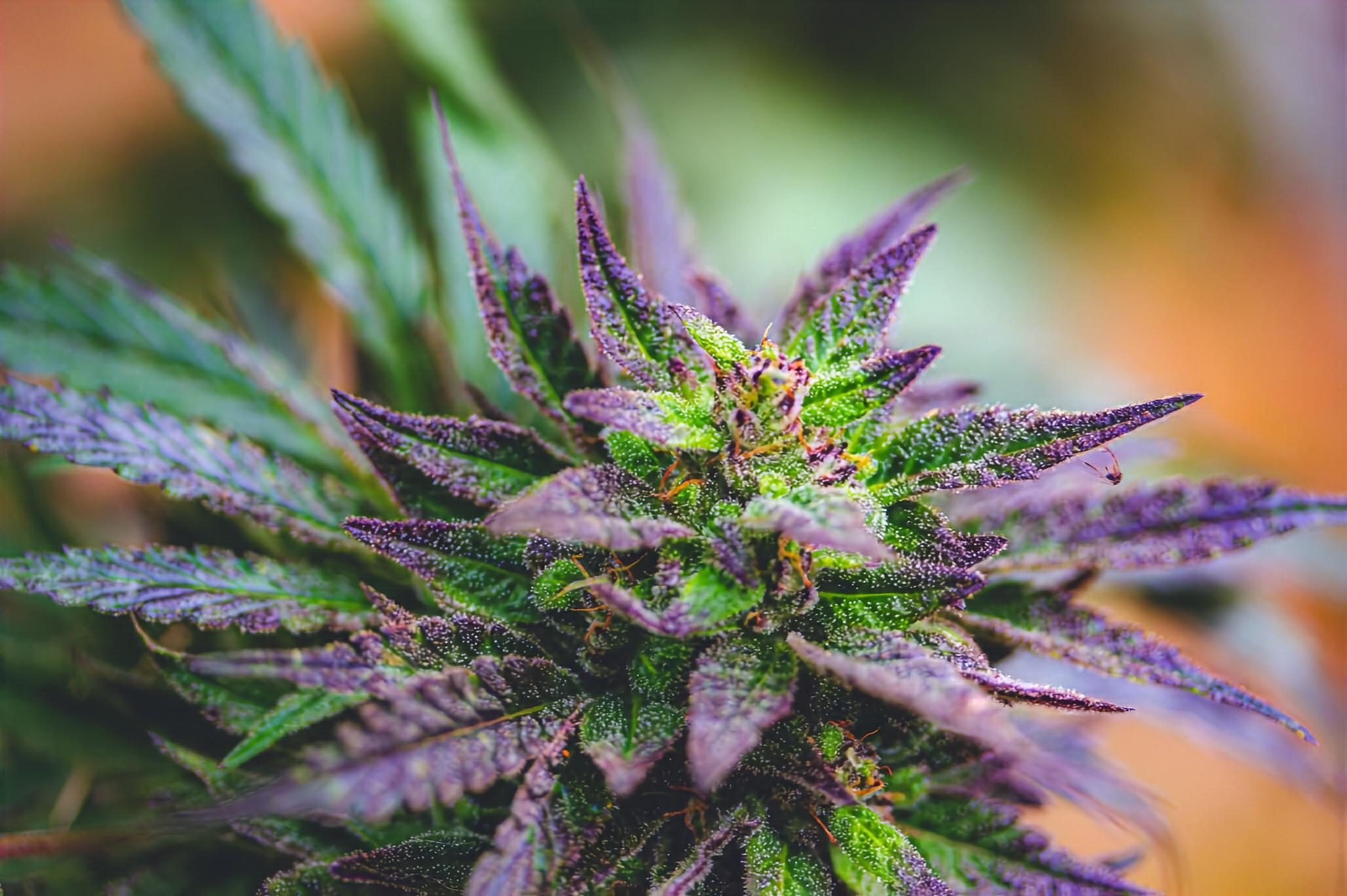}
    \caption{\ours (bpp=0.165)}
    \end{subfigure}
\end{figure}

\begin{figure}[h!]
    \begin{subfigure}[t]{0.5\linewidth}
    \centering
    \includegraphics[width=\linewidth,  height=0.35\paperheight]{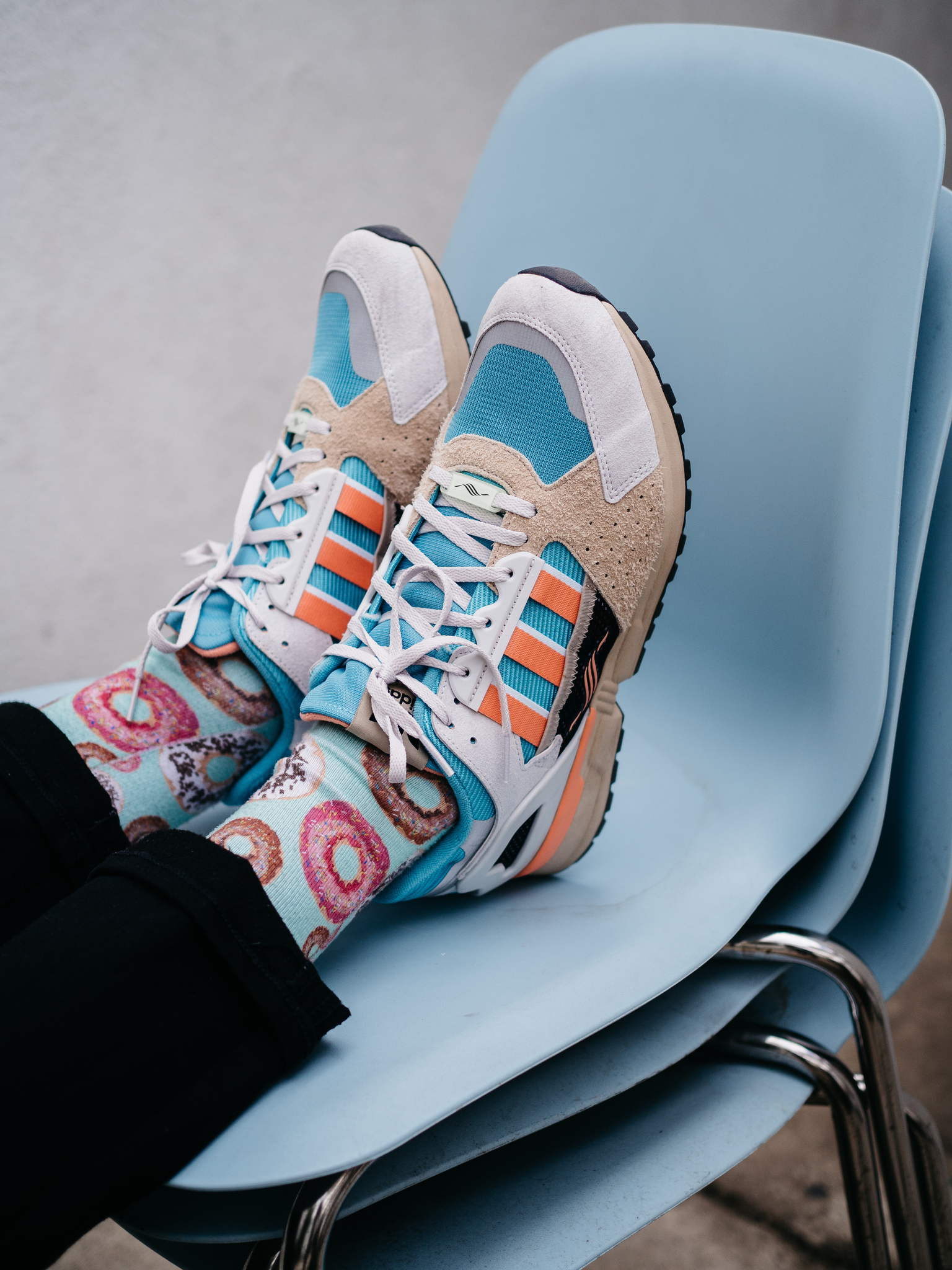}
    \caption{Original}
    \end{subfigure}
    \hfill
    \begin{subfigure}[t]{0.5\linewidth}
    \centering
    \includegraphics[width=\linewidth,  height=0.35\paperheight]{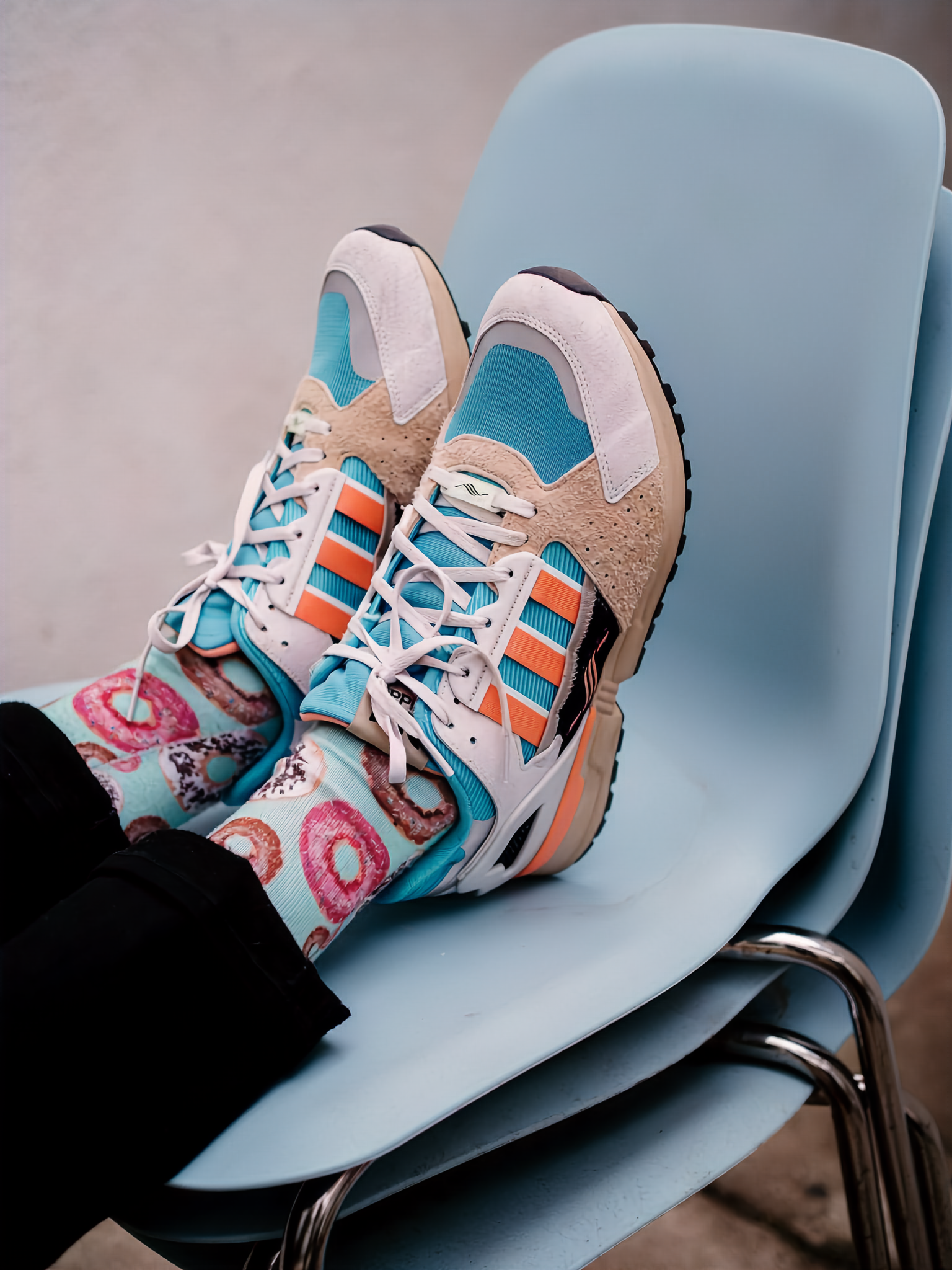}
    \caption{\ours (bpp=0.124)}
    \end{subfigure}
\end{figure}

\begin{figure}[h!]
    \begin{subfigure}[t]{0.5\linewidth}
    \centering
    \includegraphics[width=\linewidth,  height=0.35\paperheight]{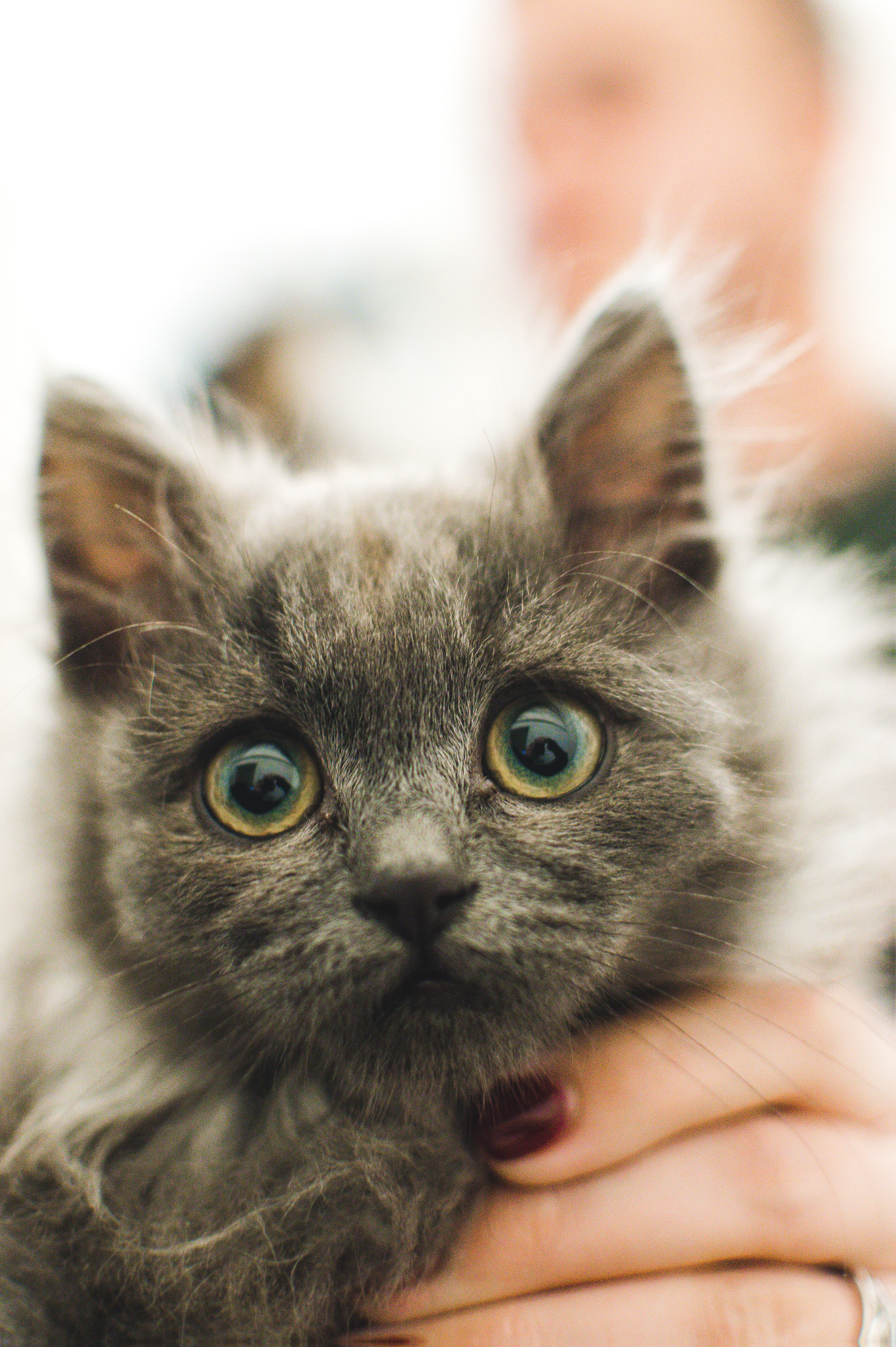}
    \caption{Original}
    \end{subfigure}
    \hfill
    \begin{subfigure}[t]{0.5\linewidth}
    \centering
    \includegraphics[width=\linewidth, height=0.35\paperheight]{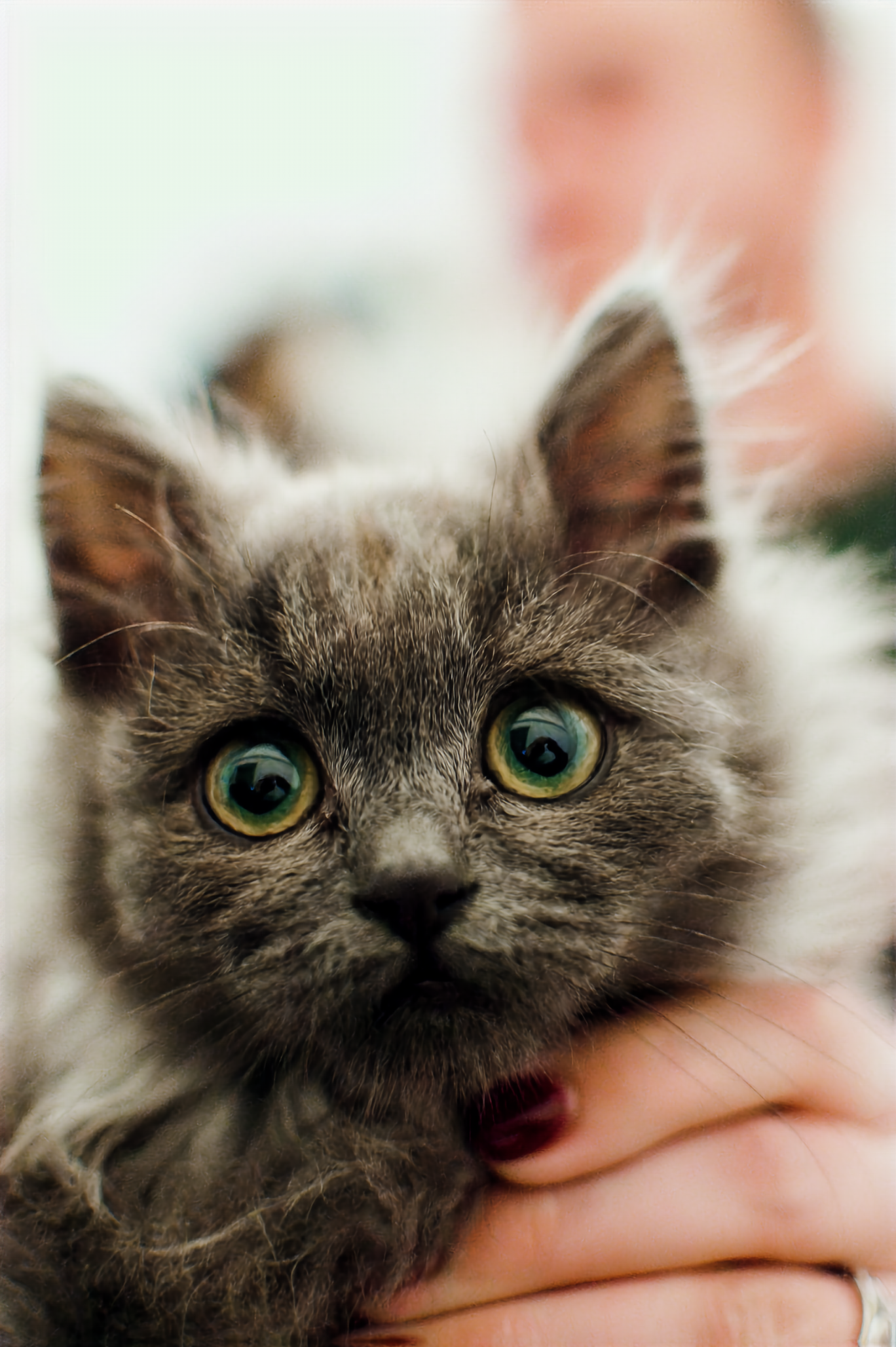}
    \caption{\ours (bpp=0.155)}
    \end{subfigure}
\end{figure}

\begin{figure}
    \begin{subfigure}[t]{0.5\linewidth}
    \centering
    \includegraphics[width=\linewidth,  height=0.35\paperheight]{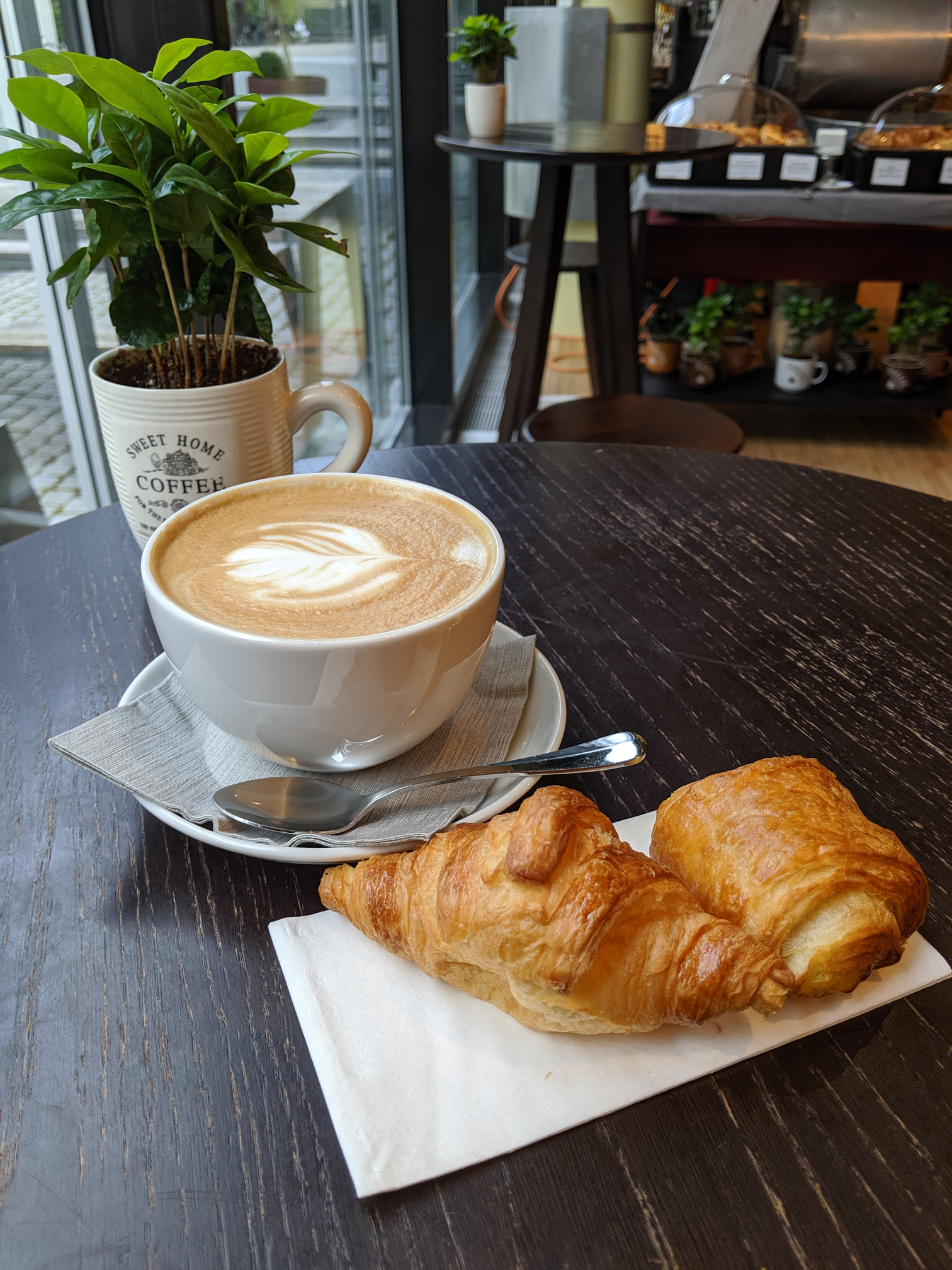}
    \caption{Original}
    \end{subfigure}
    \hfill
    \begin{subfigure}[t]{0.5\linewidth}
    \centering
    \includegraphics[width=\linewidth,  height=0.35\paperheight]{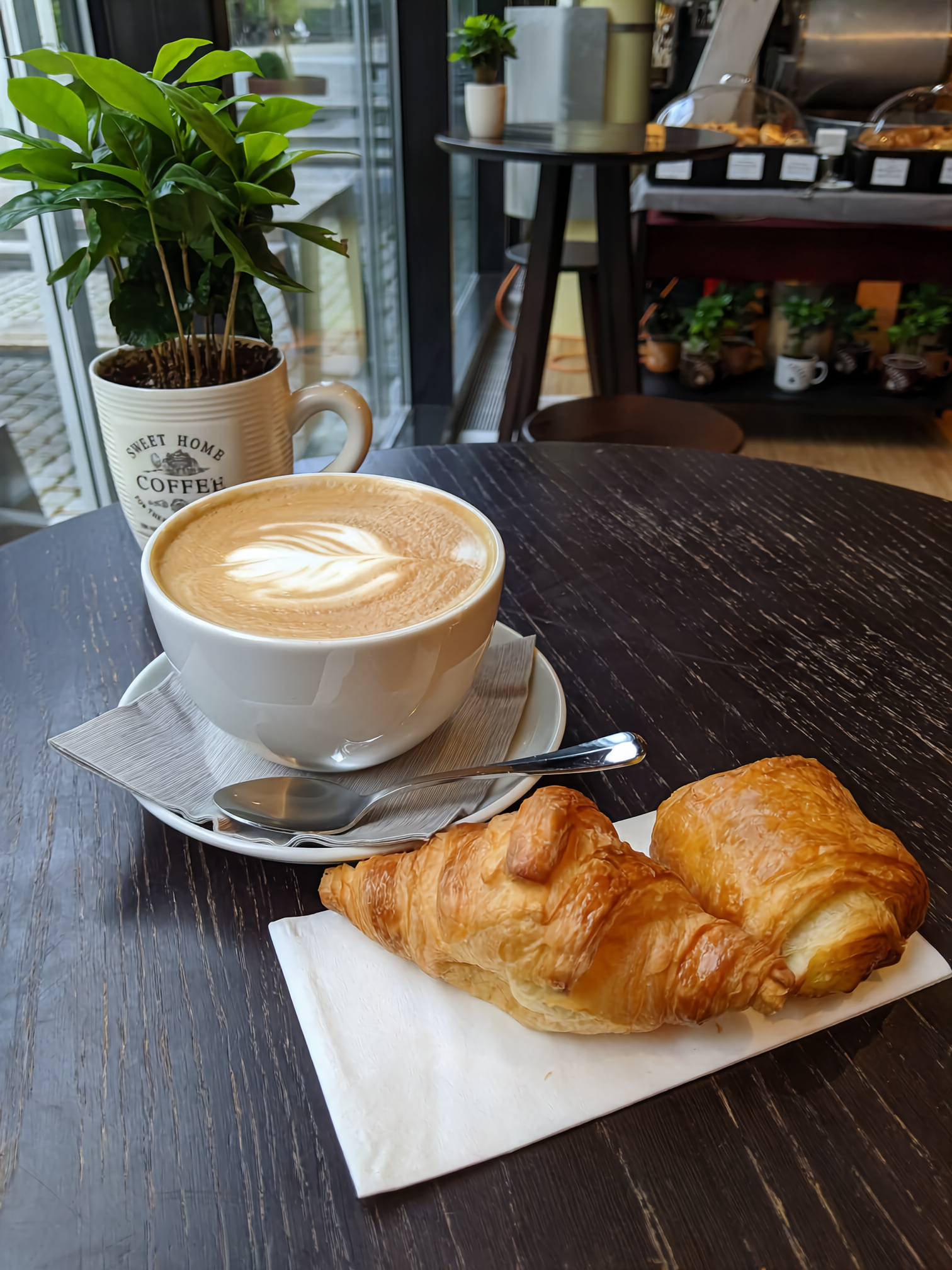}
    \caption{\ours (bpp=0.176)}
    \end{subfigure}
\end{figure}

\end{document}